%% file: main.tex
\documentclass{article}
\usepackage{CJKutf8}

\usepackage{PRIMEarxiv}
\usepackage{array}      
\usepackage{booktabs}   
\usepackage[utf8]{inputenc} 
\usepackage[T1]{fontenc}    
\usepackage{hyperref}       
\usepackage{url}            
\usepackage{booktabs}       
\usepackage{amsfonts}       
\usepackage{amsmath}        
\usepackage{amssymb}        
\usepackage{amsthm}         
\usepackage{nicefrac}       
\usepackage{amsbsy}
\usepackage{enumitem}       
\usepackage{mathrsfs}       
\usepackage{microtype}      
\usepackage{lipsum}
\usepackage{fancyhdr}       
\usepackage{graphicx}       
\graphicspath{{media/}}     
\usepackage{makecell}
\usepackage[table]{xcolor}
\usepackage{subcaption}
\usepackage{comment}
\usepackage{float}
\usepackage{multirow}
\usepackage{listings}
\usepackage{color}
\usepackage{colortbl}

\usepackage{graphicx} 
\usepackage{booktabs}

\newcommand\YAMLcolonstyle{\color{red}\mdseries}
\newcommand\YAMLkeystyle{\color{black}\bfseries}
\newcommand\YAMLvaluestyle{\color{blue}\mdseries}
\newcommand{\rotatehead}[1]{\rotatebox{15}{\textbf{#1}}}

\graphicspath{{media/}{figures/}}

\newtheorem{theorem}{Theorem}[section]

\theoremstyle{definition}
\newtheorem{definition}[theorem]{Definition}
\theoremstyle{remark}

\makeatletter

\newcommand\language@yaml{yaml}

\expandafter\expandafter\expandafter\lstdefinelanguage
\expandafter{\language@yaml}
{
  keywords={true,false,null,y,n},
  keywordstyle=\color{darkgray}\bfseries,
  basicstyle=\YAMLkeystyle,                                 
  sensitive=false,
  comment=[l]{\#},
  morecomment=[s]{/*}{*/},
  commentstyle=\color{purple}\ttfamily,
  stringstyle=\YAMLvaluestyle\ttfamily,
  moredelim=[l][\color{orange}]{\&},
  moredelim=[l][\color{magenta}]{*},
  moredelim=**[il][\YAMLcolonstyle{:}\YAMLvaluestyle]{:},   
  morestring=[b]',
  morestring=[b]",
  literate =    {---}{{\ProcessThreeDashes}}3
                {>}{{\textcolor{red}\textgreater}}1     
                {|}{{\textcolor{red}\textbar}}1 
                {\ -\ }{{\mdseries\ -\ }}3,
}

\title{JoyAI-LLM Flash: Advancing Mid-Scale LLMs with Token Efficiency}

\author{
\large JD.com
}

\begin{document}
\maketitle

\begin{abstract}

  We introduce JoyAI-LLM Flash, an efficient Mixture-of-Experts
  (MoE) language model designed to redefine the trade-off between
  strong performance and token efficiency in the sub-50B
  parameter regime. JoyAI-LLM Flash is pretrained on a massive
  corpus of 20 trillion tokens and further optimized through a
  rigorous post-training pipeline, including supervised
  fine-tuning (SFT), Direct Preference Optimization (DPO), and
  large-scale reinforcement learning (RL) across diverse
  environments. To improve token efficiency, JoyAI-LLM Flash
  strategically balances \emph{thinking} and \emph{non-thinking}
  cognitive modes and introduces FiberPO, a novel RL algorithm
  inspired by fibration theory that decomposes trust-region
  maintenance into global and local components, providing unified
  multi-scale stability control for LLM policy optimization. To
  enhance architectural sparsity, the model comprises 48B total
  parameters while activating only 2.7B parameters per forward
  pass, achieving a substantially higher sparsity ratio than
  contemporary industry leading models of comparable scale. To
  further improve inference throughput, we adopt a joint
  training–inference co-design that incorporates dense
  Multi-Token Prediction (MTP) and Quantization-Aware Training
  (QAT). We release the checkpoints for both JoyAI-LLM-48B-A3B
  Base and its post-trained variants on Hugging Face to support
  the open-source community.

\end{abstract}

\section{Introduction}
The development of highly capable Large Language Models (LLMs) is
increasingly constrained by two intertwined challenges: poor
token efficiency and high computational cost
\cite{du2026ockbenchmeasuringefficiencyllm}. During inference,
many models consume an excessive number of tokens to produce
accurate outputs. Although scaling test-time computation
\cite{snell2024scalingllmtesttimecompute} has historically
yielded substantial performance gains, there is a growing need to
fundamentally rethink intelligence from the perspective of
efficiency.

\begin{figure}
  \centering
  \includegraphics[width=0.90\textwidth]{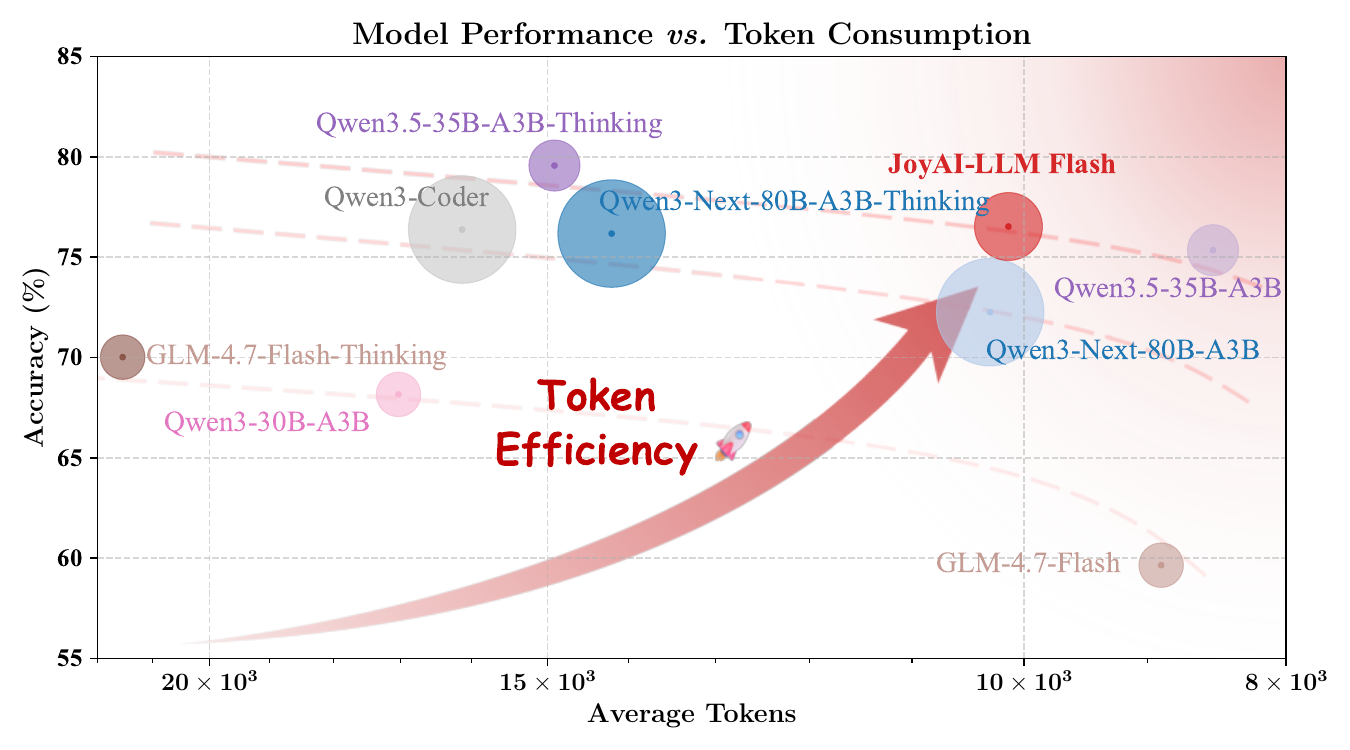}
  \caption{Model performance \emph{vs.} token consumption across different middle-scale LLMs. The accuracy and token consumption averaged across eighteen benchmarks used in post-training evaluation (Table~\ref{tab:instruct_eval}) are illustrated, where the upper-right region indicates more token-efficient models. Bubble size represents the model parameter count.}
  \label{fig:fig0}
\end{figure}

We introduce JoyAI-LLM Flash, an instruct language
model~\cite{liu2024deepseekv2} with chat, short chain-of-thought
(sCoT), and agentic capabilities. Built on a sparse
Mixture-of-Experts (MoE) architecture, JoyAI-LLM Flash
substantially advances both throughput and performance at
inference time by activating only a small fraction of its
parameters in each forward pass. Specifically, JoyAI-LLM Flash
adopts a pure attention-based architecture with a learned MLP
router that activates 8 out of 256 experts, along with 1 shared
expert. The model comprises 48B total parameters, of which only
2.7B are activated per forward pass (or 3.2B including
embeddings). As shown in Figure~\ref{fig:fig0}, JoyAI-LLM Flash
achieves competitive or superior token-efficiency performance
compared with other state-of-the-art models of similar scale. The
figure reports the average accuracy and token consumption across
eighteen benchmarks used in post-training evaluation, where
models in the upper-right region are more token-efficient.
Furthermore, under the 8K-input/16K-output setting, JoyAI-LLM
Flash achieves 1.45$\times$ and 1.07$\times$ speedups over the
pure attention-based models GLM-4.7-Flash
\cite{5team2025glm45agenticreasoningcoding} and Qwen3-30B-A3B
\cite{qwen3}, respectively. In terms of multi-token prediction
(MTP) efficiency, defined as the inference speedup of the MTP
model over its non-MTP counterpart, JoyAI-LLM Flash achieves a
1.87$\times$ speedup, surpassing the hybrid-attention models
Qwen3.5-35B-A3B \cite{qwen3.5} (1.61$\times$) and Step-3.5-Flash
\cite{huang2026step35flashopen} (1.39$\times$). We open-source
both the base and chat model weights in multiple quantization
formats.

The base model of JoyAI-LLM Flash was pretrained on an extensive
text-only corpus of over 20 trillion tokens, employing a
Warmup-Constant-Cosine-Decay learning rate schedule. To maximize
token utilization and incrementally build model capabilities, we
divide the pretraining curriculum into four stages:
\begin{itemize}
    \item Foundational Phase: Exposing the model to diverse tokens to build general linguistic capabilities.
    \item Code-Math-Enhancement Phase: Processing tokens with a significantly upweighted proportion of code and math data.
    \item Mid-Training Phase: Focusing on ultra-high-quality tokens to refine reasoning and alignment.
    \item Long-Context Phase: Utilizing nature long context tokens specifically engineered to extend the context window to 128K.
\end{itemize}
Empirically, JoyAI-LLM Flash Base achieves the competitive
efficacy and efficiency for its size class across general
knowledge, math, code, and comprehensive understanding
evaluations.

Following pretraining, we implemented a rigorous post-training
pipeline designed not only to align the model with human intent
and enhance its autonomy, but also to fundamentally optimize
token efficiency. The pipeline starts with heavily supervised
fine-tuning (SFT) on a diverse set of high-quality traces, which
strategically balance "thinking" and "non-thinking" cognitive
modes, followed by Direct Preference Optimization (DPO) to refine
responses and ensure robust human preference alignment. To
further advance its reasoning and agentic problem-solving skills,
we perform large-scale Reinforcement Learning (RL) across diverse
environments. Inspired by the algebraic concept of
\emph{fibration}, we introduce a novel RL algorithm, FiberPO,
that decomposes trust-region maintenance into global and local
components, providing unified multi-scale stability control for
LLM policy optimization. Together with these post-training
advances, JoyAI-LLM Flash emerges as a powerful foundation model,
establishing a new baseline for strong performance, token
efficiency, and computational efficiency within the sub-50B
parameter regime.

We also quantized JoyAI-LLM Flash from bfloat16 to FP8, INT8, FP4 and GGUF. Along with this report, we are releasing the model as follows:

\paragraph{Checkpoints}
\begin{itemize}
  \item \href{https://huggingface.co/jdopensource/JoyAI-LLM-Flash-Base}{JoyAI-LLM Flash Base}~\includegraphics[height=0.9em]{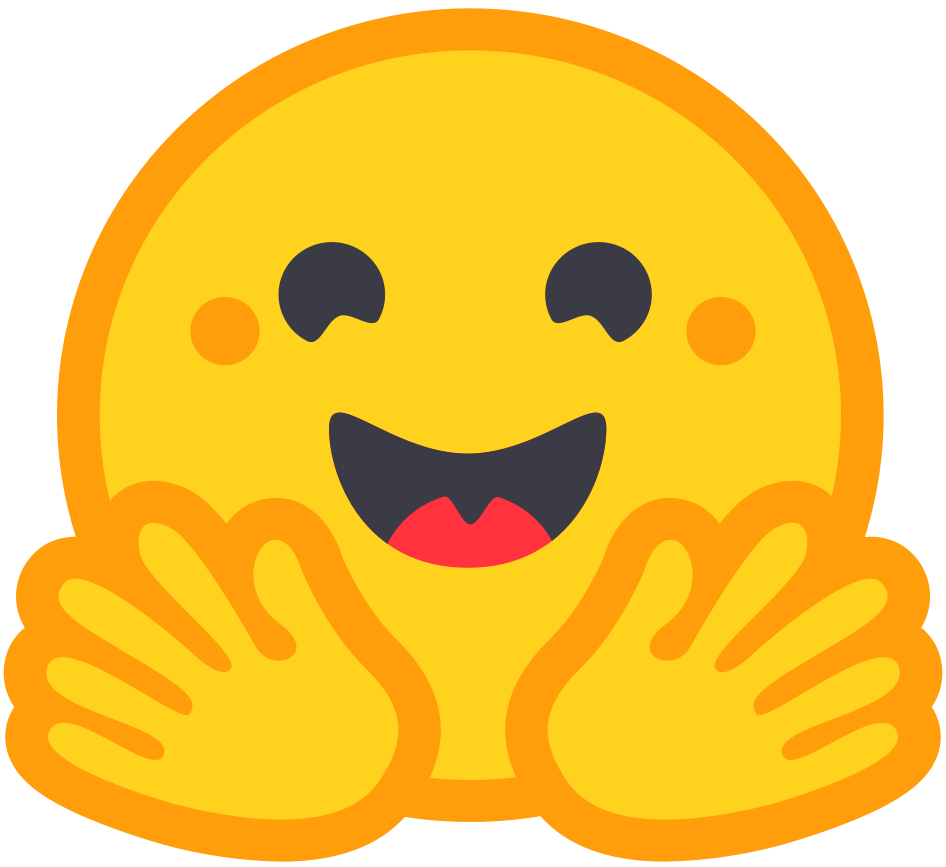}: the pre-trained base model
  \item \href{https://huggingface.co/jdopensource/JoyAI-LLM-Flash}{JoyAI-LLM Flash BF16}~\includegraphics[height=0.9em]{figures/huggingface-color.png}: the post-trained model
  \item \href{https://huggingface.co/jdopensource/JoyAI-LLM-Flash-FP8}{JoyAI-LLM Flash FP8}~\includegraphics[height=0.9em]{figures/huggingface-color.png}: the post-trained model quantized with the FP8 format delivering an excellent trade-off between performance and efficiency
  \item \href{https://huggingface.co/jdopensource/JoyAI-LLM-Flash-INT8}{JoyAI-LLM Flash INT8}~\includegraphics[height=0.9em]{figures/huggingface-color.png}: the post-trained model quantized with the INT8 format achieving an optimal trade-off between performance and efficiency and compatible with some AI accelerators
  \item \href{https://huggingface.co/jdopensource/JoyAI-LLM-Flash-INT4}{JoyAI-LLM Flash INT4}~\includegraphics[height=0.9em]{figures/huggingface-color.png}: the post-trained model quantized with the INT4 format serving as an ultra-compact variant tailored for environments with extremely restricted VRAM, such as consumer-level processors
  \item \href{https://huggingface.co/jdopensource/JoyAI-LLM-Flash-GGUF}{JoyAI-LLM Flash GGUF}~\includegraphics[height=0.9em]{figures/huggingface-color.png}: the post-trained model quantized with the GGUF ultra-low-bit format delivering broad compatibility across personal computers
\end{itemize}


The rest of this report is arranged as follows. Section
\ref{sec:pretraining} describes the pretraining process of
JoyAI-LLM Flash. Section \ref{sec:post-training} describes the
post-training process, and Section \ref{sec:inference} shows the
inference technology used in JoyAI-LLM Flash. Section
\ref{sec:conclusion} concludes the report and presents future
work.

\section{Pretraining}
\label{sec:pretraining}

In this section, we introduce the key features of JoyAI-LLM Flash Base, including its architecture, hyperparameters, and pretraining data. We also demonstrate that JoyAI-LLM Flash Base achieves competitive results to 
the state-of-the-art models.

\begin{table}[t]
    \centering
    \definecolor{graybg}{gray}{0.92} 
    \renewcommand{\arraystretch}{1.25} 
    \caption{Detailed architectural configurations of JoyAI-LLM Flash.}
    \begin{tabular}{l c} 
        \toprule
        \textbf{Hyperparameter} & \textbf{JoyAI-LLM Flash 48B-A3B} \\
        \midrule
        
        \rowcolor{graybg} \multicolumn{2}{c}{\textbf{General Settings}} \\
        Total Layers ($N_{\text{layers}}$) & 40 \\
        Dense Layers & 1 \\
        Hidden Dimension ($d_{\text{model}}$) & 2048 \\
        Vocabulary Size ($|V|$) & 129K \\
        Max Context Length & 128K \\
        Activation Function & SwiGLU \\
        
        \rowcolor{graybg} \multicolumn{2}{c}{\textbf{Attention Mechanism}} \\
        Attention Type & MLA \\
        Attention Heads ($n_h$) & 32 \\
        QK Non-RoPE Dimension ($d_{\text{nope}}$) & 64 \\
        QK RoPE Dimension ($d_{\text{rope}}$) & 128 \\
        Value Dimension ($d_v$) & 128 \\
        
        \rowcolor{graybg} \multicolumn{2}{c}{\textbf{Mixture-of-Experts}} \\
        Total Routed Experts ($N_r$) & 256 \\
        Activated Experts ($K$) & 8 \\
        Shared Experts ($N_s$) & 1 \\
        Expert Intermediate Size ($d_e$) & 768 \\
        
        \bottomrule
    \end{tabular}
    \vspace{5pt}
    \label{tab:main_arch}
\end{table}

\subsection{Model Architecture}
As summarized in Table \ref{tab:main_arch}, JoyAI-LLM Flash is a Mixture-of-Experts (MoE) model with 48.9B total parameters, of which 3.28B are activated per token. Its micro-architecture draws inspiration from DeepSeek-V3 \cite{liu2024deepseek} and Kimi-K2 \cite{team2025kimi}, utilizing Multi-head Latent Attention (MLA) \cite{liu2024deepseekv2} with hidden dimensions of 2048 and 768, respectively. The model incorporates standard components such as RMSNorm \cite{zhang2019root} for layer normalization, RoPE \cite{su2024roformer} for positional encoding, and SwiGLU \cite{dauphin2017language} activation within the feed-forward blocks.

In terms of macro-architecture, JoyAI-LLM Flash consists of 40 Transformer layers. The first layer utilizes a standard dense feed-forward network, while the remaining 39 layers are sparse MoE layers. The MoE module employs a fine-grained architecture with 256 total experts. For each input token, the model activates a total of nine experts: eight routed experts are dynamically selected via a Top-8 gating mechanism, and a single dedicated shared expert is always activated to capture common knowledge. To ensure numerical stability, the gating scores are computed using a sigmoid function and executed in FP32 precision. Additionally, we implement an auxiliary-loss-free load-balancing strategy \cite{wang2024auxiliary} to maintain optimal utilization across the expert pool. 


\textbf{Muon Optimizer}. To maximize sample efficiency and accelerate convergence, we employ the Muon optimizer \cite{team2025kimi, jordan6muon}. Unlike standard Adam, which relies on element-wise scaling, Muon optimizes parameters by leveraging matrix orthogonalization, effectively performing a form of Newton-style update on the spectral norm of the gradients. Previous studies \cite{team2025kimi, jordan6muon} have demonstrated that this approach significantly accelerates model convergence compared to Adam-based optimizers. Beyond these advantages, our empirical results reveal that Muon substantially enhances training robustness. During our experiments, training sessions utilizing Adam were frequently plagued by several loss spikes, which required manual intervention or careful adjustment of learning rates. In contrast, training with Muon remained exceptionally stable, with no significant numerical anomalies observed.


\textbf{Multi-Token Prediction}. We append a lightweight, single-layer dense Multi-Token Prediction (MTP) head to jointly optimize training and inference \cite{liu2024deepseek, xiao2026mimo}. During pre-training, this module enriches the learning signal, enabling the model to capture multi-step dependencies for improved data efficiency. During inference, it natively enables speculative decoding. By predicting multiple future tokens in parallel, this mechanism overcomes the latency bottleneck of standard autoregressive decoding and accelerates generation without requiring an external draft model.

\subsection{Infrastructure}

The JoyAI-LLM Flash training system is built upon a highly optimized extension of the Megatron-Core \cite{shoeybi2019megatron} framework. Along with foundational parallelization strategies—including Data Parallelism (DP), Tensor Parallelism (TP) \cite{shoeybi2019megatron}, Sequence Parallelism (SP) \cite{shoeybi2019megatron}, and the 1F1B Pipeline Parallelism (PP) \cite{huang2019gpipe, harlap1806pipedream, harlap1806pipedream, lamy2023breadth, liu2023hanayo, narayanan2021efficient, qi2023zero, combine-1f1b} schedule, we have introduced several architectural enhancements to maximize throughput and computational efficiency. Our specific configuration employs 2-way Pipeline Parallelism, 8-way Expert Parallelism (EP) \cite{lepikhin2020gshard} spanning two nodes, and ZeRO-1 Data Parallelism \cite{rajbhandari2020zero}. To further accelerate core operations, we integrate FlashAttention-3 \cite{NEURIPS2024_7ede97c3} for high-performance attention kernels and utilize the DeepEP \cite{liu2024deepseek} library to minimize latency during token dispatch and combination within the MoE layers. Additionally, by leveraging distributed asynchronous checkpointing, we reduce loading times from 15 minutes to 30 seconds, ensuring the model can recover and resume training in less than a minute. We also implement a packing training strategy utilizing block-diagonal masks to isolate unrelated samples and preserve strict causal boundaries. Compared to the conventional full lower-triangular masking approach, this method accelerates the attention forward and backward passes by 50\% and 20\%, respectively.

\subsection{Pretraining Data} 

In this section, we detail the composition and processing pipeline of our pretraining corpus. Our model was trained on a total of 20.7 trillion high-quality tokens. The dataset is curated from four main sources: diverse web crawls, reasoning-intensive code repositories, high-fidelity PDF documents, and large-scale synthetic data. The composition is designed to balance broad general knowledge with deep reasoning capabilities and domain-specific expertise.

\subsubsection{Web Data Pipeline}
We processed Common Crawl data up to October 2025 using a high-efficiency pipeline:

\textbf{Text Extraction.} To achieve high-quality content extraction, we process WARC files using the Trafilatura library, which more effectively removes boilerplate and menu text while filtering out HTML artifacts to extract the core semantic text.

\textbf{Rule-based Cleaning.} Our data refinement process utilizes the \textit{Datatrove} framework \cite{penedo2024datatrove}, integrated with several customized modules for high-precision filtering:
\begin{itemize}
\item \textbf{Standard Filtering:} We employ URL filtering to block known malicious domains and a \textit{fastText} classifier to retain only high-confidence English and Chinese documents. Content quality is further ensured through quality and heuristic repetition filters.
\item \textbf{Privacy Preservation:} The PII (Personally Identifiable Information) detection logic was significantly expanded to cover a more comprehensive set of global identity markers, providing robust anonymization.
\item \textbf{Optimized Decontamination:} To address the issue of excessive data removal during decontamination, we introduced an n-gram whitelisting mechanism. This refinement reduces the probability of "false deletions", ensuring that only true evaluation overlaps are removed.
\item \textbf{Semantic Safety Classifier:} A dedicated BERT-based sensitive content classifier was added to our pipeline. This model performs deep semantic analysis to detect policy-prohibited data, ensuring the final dataset aligns with safety and ethical guidelines.
\end{itemize}

\textbf{Deduplication.} To mitigate redundancy, we developed a distributed deduplication pipeline based on MinHash-LSH\cite{code-lsh, code-minhash} on a Ray-based distributed cluster. Our approach involves decomposing documents into 7-gram shingles, followed by the generation of compact 128-permutation MinHash signatures. We employ a Jaccard similarity threshold of 0.9 to identify near-duplicate candidates. These candidates are subsequently clustered using a parallelized Union-Find algorithm, ensuring that only a single canonical representative from each equivalence class is retained in the final corpus.

\textbf{Model-based Filtering.} To address the limitations of static rules in capturing nuanced quality, we fine-tuned Qwen \cite{qwen3} model series to create two specialized filtering models:

\begin{itemize}
\item \textbf{Line-level Noise Filter:} In filtered web crawled data, we observed persistent noise such as embedded advertisements, navigation bars, and templated boilerplate. We trained a lightweight classifier to evaluate each line, removing non-narrative content while preserving the semantic integrity of the document.
\item \textbf{Multi-dimensional Scoring and Classification:} To ensure the highest data quality, we evaluates every document across several key metrics: factual accuracy, linguistic coherence, information density, grammatical correctness, thematic depth, web noise, safety and multi-topic identification. 
We only retained documents categorized as "high-quality" based on the integrated scores. This strict filtering significantly improved the model's learning efficiency.
\end{itemize}

\subsubsection{Code Data Pipeline}

\textbf{Rule-Based Cleaning Pipeline}. Our raw corpus primarily comes from The Stack v2\cite{code-stackv2} and large-scale GitHub code extraction. We follow a multi-stage cleaning workflow: rule-based filtering to remove obvious noise, model-based scoring for second-stage selection and stratification, and deduplication to reduce redundancy. Generic quality signals capture repetition (high-order n-gram duplication and duplicate-line ratios), length and scale (character/line counts, file size, extreme line lengths), character composition (ratios of alphabetic/digit/whitespace characters, hex-like fragments, hyperlinks/HTML tags), and suspicious patterns such as autogenerated or encoded content. Language-specific signals further characterize structural and semantic parseability (e.g., AST availability), function-to-line ratios, test-file patterns, preprocessing-directive density, and excessive trivial statements (e.g., \texttt{print}, \texttt{assert}, \texttt{pass}). Guided by downstream scoring, we relax overly aggressive heuristics by moving them from hard filtering to the scoring stage, improving the precision--coverage trade-off.

\textbf{Model-Based Quality Scoring}. To reduce the cost of per-sample LLM evaluation, we train a lightweight regression scorer to approximate a large-capacity judge model. We use Qwen2.5-3B-Instruct\cite{qwen2025qwen25technicalreport} as the backbone and target sequences shorter than 32k tokens. Labels are produced by Qwen2.5-Coder-32B-Instruct\cite{code-qwen25coder}, which scores each sample 10 times; we take the minimum score to mitigate stochasticity. We define medium-quality data as scores in (2.5, 6) and high-quality data as scores greater than 6.

\textbf{MinHash-LSH Deduplication}. To remove exact and near-duplicate code, we reuse the MinHash-LSH based Ray deduplication scheme introduced in our data cleaning pipeline. In practice, most code duplicates we eliminate are exact file-level copies rather than merely similar variants. 


\textbf{Long-Context Code Construction}. For long-context code data, we construct 64k and 128k token sequences by concatenating longer QA pairs and repository-level code that have already passed the preceding filtering and deduplication stages. For repositories, we build lightweight language-specific dependency graphs over modules and files, and within each connected component we apply a topological ordering routine to obtain file sequences that respect dependency directions from lower-level to higher-level modules—that is, an ordering in which every dependency appears before any module that depends on it; for repository-level data we explicitly split the mixture so that roughly half of the sequences follow this topological order and the other half use a random file ordering, exposing the model both to realistic project layouts and to more diverse context permutations.

\textbf{Code Rewriting and Composition}. For synthetic code data, we adopt a single-pass rewriting strategy inspired by SwallowCode\cite{code-swallowcode} and OLMo3\cite{code-olmo3} (SGCR+SCOR). We start from the deduplicated and filtered GitHub corpus and a curated top-20-language subset as seeds, and ask an LLM to rewrite functions, files, and small multi-file snippets into more instruction- and documentation-like forms while preserving semantic intent; rewritten outputs are again deduplicated against both the seed pool and the original source files, routed through the same rule filtering and model scoring pipeline, and we retain samples with scores > 7, most of which empirically fall into the high-quality range (typically scoring above 8).

For QA-style code data, we primarily rely on Nemotron-Competitive-Programming-v1\cite{nvidia2025nemotron3nanoopen} as seed problems and solutions, and use the DeepSeek V3.2\cite{deepseekai2025deepseekv32} model to produce paired “thinking” and “no-thinking” variants; these rewritten QA examples are then treated as code–QA compositions and subjected to the same deduplication and quality filtering stack as the rest of the corpus.

\subsubsection{PDF Parsing and Knowledge Extraction}
To mitigate the scarcity of high-quality, specialized content in open-web corpora, we curated a massive dataset comprising tens of millions of PDF documents. This collection prioritizes domain-specific knowledge often underrepresented in general web text, including STEM, Medicine, Social Sciences, Education, Humanities, and Law. The processing pipeline follows a rigorous workflow similar to our web text pipeline, with specialized enhancements for the PDF format:

\begin{itemize}
\item \textbf{Text Extraction:} We leverage MinerU \cite{niu2025mineru25decoupledvisionlanguagemodel} and DeepSeek-OCR \cite{wei2025deepseek} to perform high-fidelity document parsing. This ensures the precise recovery of complex mathematical formulas, tables, and hierarchical structures.

\item \textbf{Filtering and Deduplication:} We implement a series of heuristic rules to rectify post-extraction artifacts, such as repetition or extraction noise. Furthermore, documents are unsuitable for linguistic modeling are systematically filtered.

\item \textbf{Semantic Chunking:} Documents are partitioned into segments of approximately 4,096 tokens by utilizing double-newline delimiters as natural boundaries, we ensure that chunks respect semantic integrity, thereby avoiding arbitrary truncation during the training phase.

\item \textbf{Quality Scoring:} Finally, the extracted text undergoes a scoring and filtering process consistent with our web-scale pipeline to ensure high data quality.
\end{itemize}
Unlike standard web crawls, these documents provide the structured, professional knowledge essential for the model to acquire advanced academic and technical expertise.

\subsubsection{Large-Scale Data Synthesis}

Synthetic data plays a critical role in our data pipeline, evolving from strengthening factual knowledge in early training to eliciting advanced, multi-step reasoning and agentic behavior in later stages.

\textbf{Factual-knowledge reformulation}. We synthesize factual pre-training data via two complementary transformations. First, we apply the MAGA reformulation method \cite{hao2025reformulationpretrainingdataaugmentation} to rewrite high-quality web passages into diversified, instruction-following styles while preserving the original semantics, thereby expanding stylistic coverage and reducing template bias. Second, we perform Nemotron-CC–style QA rewriting \cite{Su2024NemotronCCTC}, converting curated Common Crawl text into question–answer pairs and short instructional exemplars that elicit explicit information retrieval and grounded responses. 

\textbf{Long-form reasoning QA synthesis}. We construct reasoning-intensive supervision through two streams. First, we use DeepSeek V3.2 \cite{deepseekai2025deepseekv32} to generate full solutions for real-world STEM problems, and retain only responses that are verified either by major voting or by matching available ground truth. Second, inspired by Nvidia’s RQA method \cite{nvidia2025nemotron3nanoopen}, we derive graduate-level “Thinking QA” from STEM papers by turning technical content into questions that require multi-step derivations, and we synthesize multiple independent reasoning paths per question to encourage robust, self-consistent reasoning rather than pattern imitation. Overall, during the mid-training stage, we increase the proportion of synthetic data to above 60\% of total tokens to explicitly prioritize advanced reasoning.

\begin{figure}
  \centering
  \includegraphics[width=1.0\textwidth]{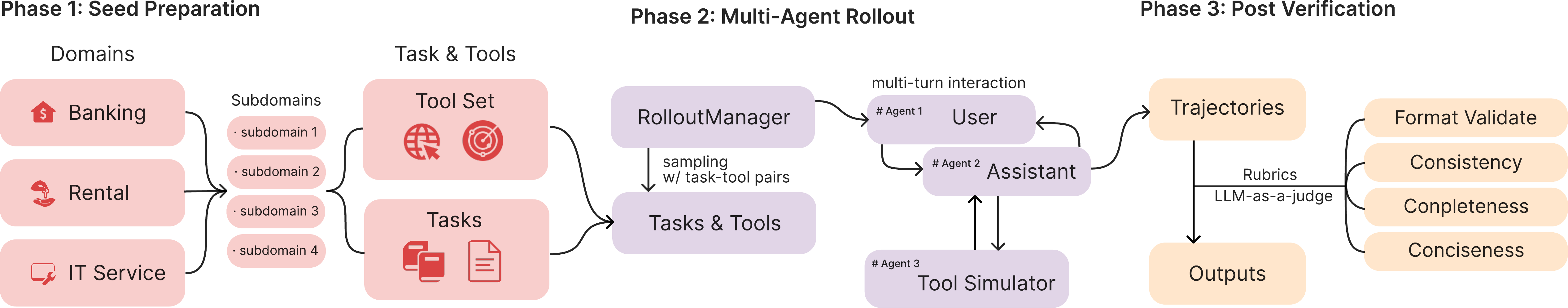}
  \caption{Agentic trajectory synthesis pipeline}
  \label{fig:agentic_traj_synth_pipeline}
\end{figure}

\textbf{Agentic trajectory synthesis}. To further augment general agentic capabilities, we complement the reasoning-centric mixture with large-scale tool-use trajectories generated through a staged execution pipeline, as illustrated in Figure \ref{fig:agentic_traj_synth_pipeline}. We first sample diverse atomic tasks across domains, then compose them into more challenging multi-intent tasks while removing repeated goals and duplicate combinations. These tasks are compiled into executable scripts and instantiated via multi-turn simulations, where GLM-4.6 \cite{5team2025glm45agenticreasoningcoding} serves as the primary agentic actor to simulate complex user--agent interactions under varied patterns. We subsequently employ an LLM-based evaluator to filter trajectories against a comprehensive set of rubrics, covering aspects such as role consistency, task completeness, and planning conciseness. 

\subsection{Training Strategy} 

We train the model using a rampup strategy, combined with a Warmup-Stable-Decay (WSD) learning rate schedule \cite{hu2024minicpm}. A sequence length of 4,096 tokens is used throughout training, except during the context extension stage.

\textbf{Stage 1 (Foundational pretraining)}. During the warmup phase, we train on 100B tokens while linearly increasing the batch size from approximately 13M to 38M tokens and ramping the learning rate to a peak of \(4.2\times 10^{-4}\). In the Stable phase, we maintain a batch size of 38M tokens and a learning rate of \(4.2\times 10^{-4}\). 

\textbf{Stage 2 (Code-Math-Enhancement pretraining)}. This phase is also regarded as the decay phase, we train with the learning rate following a cosine schedule from \(4.2\times 10^{-4}\) to \(1.4\times 10^{-4}\).

\textbf{Stage 3 (Mid-training)}. We continue training high-quality data to further refine the model. The learning rate decays from \(1.4\times 10^{-4}\) to \(4.2\times 10^{-5}\). In this stage, we enable a single-layer dense Multi-Token Prediction (MTP) \cite{liu2024deepseek, xiao2026mimo} with a loss scaling factor of 0.1.

\textbf{Stage 4 (Context Extension)}. Training proceeds in two steps, retaining the same Multi-Token Prediction configuration as Stage 2. First, we train with a 64K context window and a batch size of approximately 34M tokens, decaying the learning rate from \(4.2\times 10^{-5}\) to \(3.2\times 10^{-5}\). We then extend the context window to 128K and train with the learning rate further decaying to \(2.0\times 10^{-5}\).

\begin{figure}[htbp]
  \centering
  \begin{subfigure}[t]{0.48\textwidth}
    \centering
    \vspace{0pt} 
    \includegraphics[width=1\linewidth]{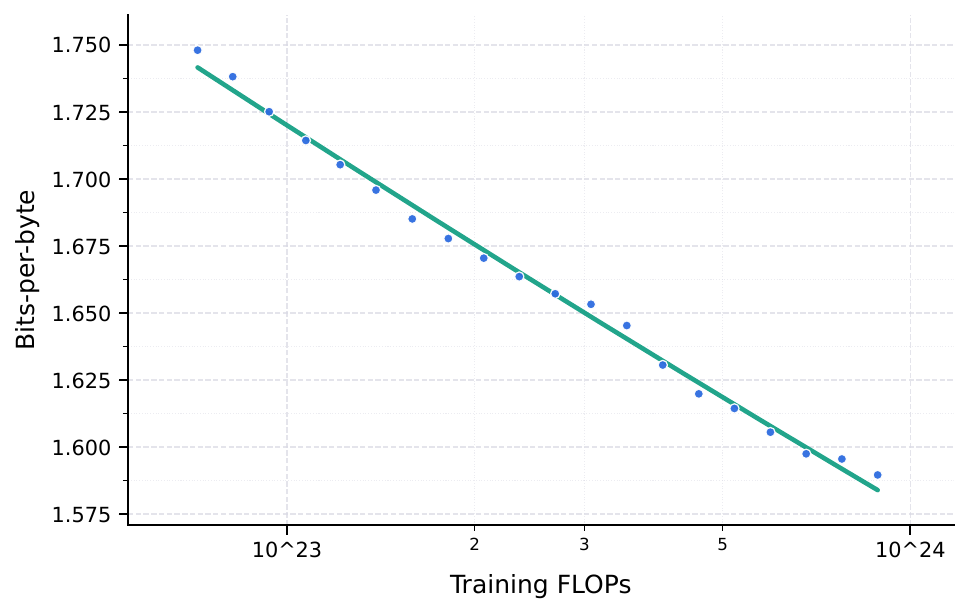}
    \caption{LOSS}
  \end{subfigure}\hfill
  \begin{subfigure}[t]{0.48\textwidth}
    \centering
    \vspace{0pt} 
    \includegraphics[width=1\linewidth]{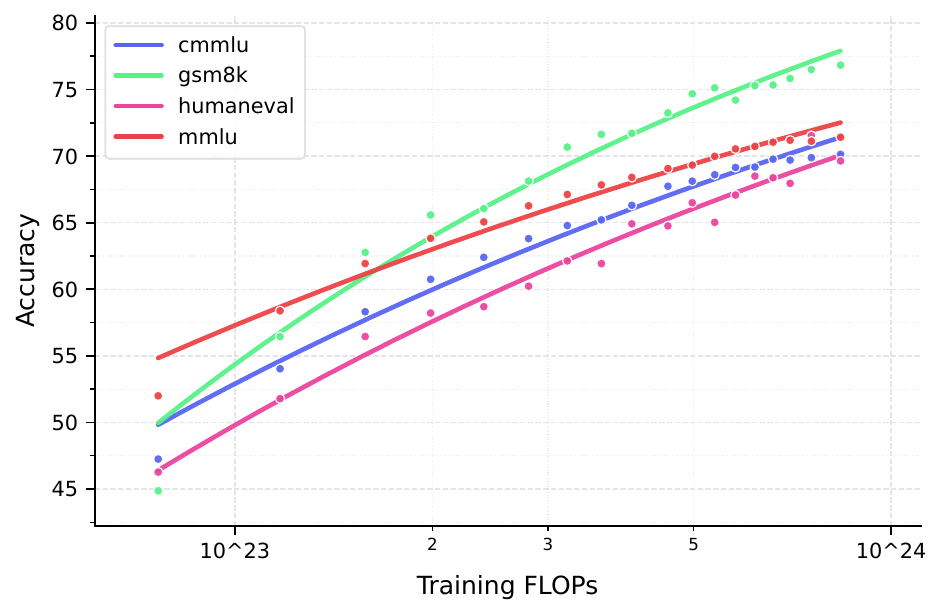}
    \caption{Benchmark}
  \end{subfigure}
  \caption{Scaling laws for JOYAI-LLM Flash. The plot illustrates the relationship between training compute and model performance. Data points represent empirical observations, while solid lines indicate the power-law fits.}
  \label{fig:scale_preformance}
\end{figure}

\textbf{Scaling laws}. During our model training process, we utilized the scaling law algorithm to guide and inform our approach. Building on previous work~\cite{scalinglaw_google, scalinglaw_google2, scalinglaw_openai}, we experimented with scaling both the model size and data volume. This strategy was crucial for anticipating the model’s training needs due to the extensive resource requirements. 
The scaling law provided predictive insights throughout the training, particularly in terms of resource allocation, as well as adjustments to training hyperparameters and data compositions. As shown in the Figure~\ref{fig:scale_preformance}, we compared the scaling law with model training loss and domain-specific benchmarks. The learning curve for training loss aligned perfectly with the scaling law’s step predictions. Although downstream tasks displayed more variability, their performance was still commendable, generally fluctuating around the scaling law trend. 
A noteworthy discovery was the use of model merging to simulate a decay in the learning rate. This approach led to more stable improvements in the performance of sub-domain tasks, aligning with the scaling law trajectory. This finding underscores the potential of model merging to optimize learning dynamics within the framework of scaling laws.

    
    
    

\subsection{Base Model Evaluation}

To evaluate the comprehensive performance of JoyAI-LLM Flash, we select Qwen3-30B-A3B-Base \cite{qwen3} and the latest Qwen3.5-35B-A3B-Base \cite{qwen3.5} as our competitive baselines. Our evaluation framework is structured around four core domains: General Knowledge, Math, Coding, and Long-Context Processing. This multidimensional evaluation thoroughly validates the model's foundational capabilities with nine benchmarks.

\begin{itemize}
\item \textbf{General Knowledge}: MMLU~\cite{hendrycks2020measuring} (5-shot, Cot), MMLU-Pro~\cite{wang2024mmlu} (5-shot, Cot), CMMLU~\cite{li2024cmmlu} (5-shot, Cot).
\item \textbf{Math}: GSM8K~\cite{cobbe2021training} (4-shot, Cot), MATH~\cite{hendrycks2021measuring} (4-shot), MATH-500~\cite{hendrycks2021measuring} (4-shot).
\item \textbf{Coding}: HumanEval~\cite{chen2021evaluating} (5-shot), LiveCodeBench~\cite{jain2024livecodebench} (v6, 2023.05-2025.04).
\item \textbf{Long-Context}: RULER~\cite{hsieh2024ruler}.
\end{itemize}
To ensure fair and reproducible comparisons, we adopt a standardized evaluation pipeline. Most benchmarks are executed with OpenCompass under greedy decoding for deterministic reporting. LiveCodeBench and RULER are evaluated using their official repositories to remain consistent with their native leaderboards. For LiveCodeBench, we use the default settings (Temperature=$0.2$, Top-$p$=$0.95$, Top-$k$=$20$, Repetition Penalty=$1.05$); for RULER, we follow the official execution protocol.

\begin{table}[htbp]
    \centering
    \definecolor{graybg}{gray}{0.9} 
    \caption{Comparison of Base Model between Qwen3-30B-A3B, Qwen3.5-35B-A3B and JoyAI-LLM Flash. Best results are marked in bold.}
    \begin{tabular}{lccc}
        \toprule
        \textbf{Task} & \textbf{Qwen3-30B-A3B-Base} & \textbf{Qwen3.5-35B-A3B-Base} & \textbf{JoyAI-LLM Flash-Base} \\
        \midrule

        \rowcolor{graybg}
        \multicolumn{4}{l}{\textbf{General Knowledge}} \\ 
        \quad MMLU       & 82.1 & \textbf{88.4} & 84.7 \\
        \quad MMLU-Pro   & 61.7 & 60.7 & \textbf{73.1} \\
        \quad CMMLU      & 83.6 & \textbf{86.1} & 83.1 \\
        \midrule 

        \rowcolor{graybg}
        \multicolumn{4}{l}{\textbf{Math}} \\
        \quad GSM8K          & 90.3 & \textbf{90.5} & 88.7 \\
        \quad MATH           & 59.6 & 56.0 & \textbf{78.1} \\
        \quad MATH-500       & 58.0 & 54.8 & \textbf{77.0} \\
        \midrule 

        \rowcolor{graybg}
        \multicolumn{4}{l}{\textbf{Coding}} \\
        \quad HumanEval       & \textbf{87.8} & 79.8 & 85.3 \\
        \quad LiveCodeBench   & 37.3 & \textbf{42.6} & 39.9 \\
        \midrule 


        \rowcolor{graybg}
        \multicolumn{4}{l}{\textbf{Long-Context}} \\
        \quad RULER (128K)            & 61.8 & \textbf{88.3} & 77.0 \\
        
        \bottomrule
    \end{tabular}%
    \vspace{10pt}
    \label{tab:main_results_lines}
    \renewcommand{\arraystretch}{1.2} 
\end{table}
Based on the results in Table \ref{tab:main_results_lines}, JoyAI-LLM Flash shows a competitive profile relative to the Qwen models. On broad general-knowledge benchmarks, it is slightly behind the strongest baseline, suggesting its factual coverage is comparable but not consistently better. On reasoning-intensive and math evaluations, it performs more strongly, indicating better robustness on harder multi-step problem solving under the reported setup. For coding, results are broadly on par with the baselines, with small differences depending on the benchmark.

\section{Post-Training} 
\label{sec:post-training}
In contrast to contemporary mid-scale models, JoyAI-LLM Flash dedicates a significantly larger proportion of its computational budget to the post-training phase. We structure this rigorous alignment pipeline into three sequential stages: Supervised Fine-Tuning (SFT), Direct Preference Optimization (DPO), and Reinforcement Learning (RL). During the SFT stage, we deliberately interleave "thinking" and "non-thinking" cognitive data mixtures. Empirical observations indicate that this hybrid training approach substantially enhances the performance of the instruct model (non-thinking model). 
Following SFT, we introduce a dedicated DPO phase to refine response quality and mitigate hallucinations. The inclusion of DPO before RL is strategically motivated by its rapid convergence, providing a highly efficient mechanism for penalizing negative or undesirable responses early in the alignment process.
Finally, building upon the DPO-aligned foundation, we apply a novel, large-scale RL algorithm designed to maximize token efficiency and further elevate the model's agentic problem-solving capabilities.

\subsection{Supervised Fine Tuning (SFT)}
We establish that the SFT phase is fundamental to realizing the comprehensive capabilities of JoyAI-LLM Flash. Rather than merely aligning output formats or following instructions, this stage is instrumental in expanding the model's knowledge and amplifying its core cognitive capacities. To this end, we implement a heavily weighted SFT protocol comprising a diverse training mixture across three distinct categories: general SFT, environment and agent learning, and tool-integrated reasoning (TIR). 
\subsubsection{General SFT}
Our general Supervised Fine-Tuning (SFT) corpus encompasses a comprehensive spectrum of domains, including mathematics, coding, tool utilization, instruction following, safety, science, Lean theorem proving, creative writing, role-playing, language and multilingual understanding. To construct this dataset, we aggregate substantial volumes of real-world and synthetic prompts, paired with high-quality responses derived from both human annotators and state-of-the-art open-source models, such as JoyAI-LLM Pro, DeepSeek-V3.2 \cite{deepseekai2025deepseekv32}, Qwen3-235B-A22B \cite{qwen3}, and GPT-OSS 120B \cite{openai2025gptoss120bgptoss20bmodel}. To maintain stringent quality standards, we employ Qwen3-30B-A3B \cite{qwen3} as a specialized filter to systematically remove low-quality queries. Notably, we eschew curriculum learning during this phase in favor of a unified training approach.

Recognizing the pivotal role that data mixture plays across both the mid-training and SFT stages, we apply a human-in-the-loop scheme \cite{omniforce} to dynamically optimize domain proportions. Specifically, we heavily weight coding and agent-centric data to constitute approximately 30\% of the mixture, followed closely by general chat and STEM domains.

To maximize computational efficiency during training, we pack sequences to a context length of 128K using a best-fit packing algorithm. This technique reduces the padding ratio to a negligible 0.01\%, ensuring that computational resources are exclusively allocated to effective tokens. To preserve strict causal boundaries and prevent cross-contamination between unrelated samples, we apply block-diagonal attention masks within the packed sequences. Overall, this sequence packing strategy yields a 1.5x improvement in training throughput compared to standard padding methodologies.

\subsubsection{Environment and Agent Learning}

\begin{figure}
  \centering
  \includegraphics[width=0.90\textwidth]{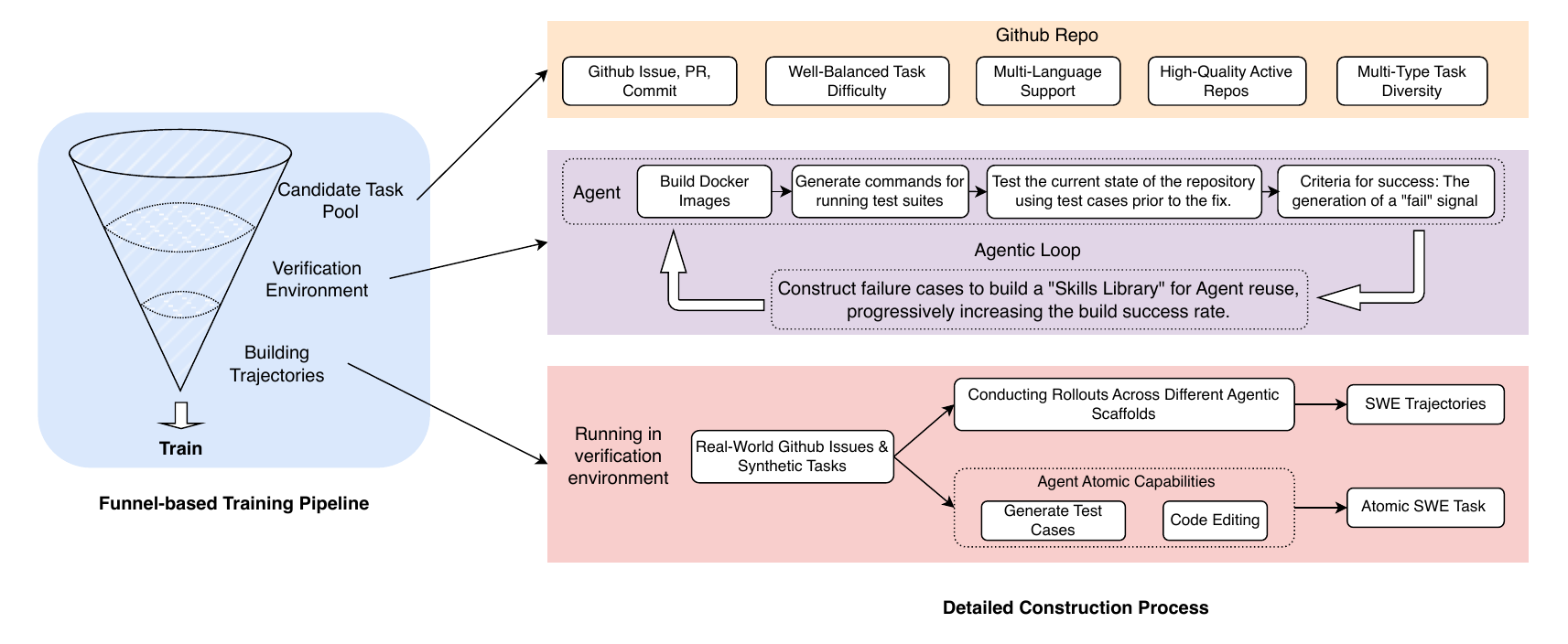}
  \caption{Verifiable Environment Pipeline}
  \label{fig:fig4}
\end{figure}
To systematically optimize Software Engineering (SWE) tasks, as illustrated in Figure 4, we modeled the workflow as a conversion funnel, analogous to the traffic funnels used in internet marketing. From the initial mining of tasks on GitHub to the final rollout of agent trajectories, each successive stage involves a natural "conversion loss," where the pool of viable tasks diminishes. By monitoring each stage of this pipeline through the lens of a funnel model, we can precisely analyze bottlenecks and guide optimization efforts to improve the overall conversion rate of usable tasks. The pipeline is structured into three primary phases:
\begin{itemize}
\item \textbf{Candidate Task Mining} The initial stage focuses on excavating raw candidate tasks from GitHub. This involves filtering repositories and issues based on specific heuristics to ensure the tasks are substantive and representative of real-world engineering challenges.
\item \textbf{Verifiable Environment Construction} Following mining, we transition to the environment setup phase. Here, we transform raw tasks into reproducible environments. The objective is to ensure that each task has a functional "ground truth" (e.g., passing/failing tests) that can be used for automated verification and subsequent Reinforcement Learning (RL).
\item \textbf{Trajectory Generation and Cold Start} The final stage involves rolling out interaction trajectories using an agent framework. These successful trajectories (pass the test cases) serve as high-quality data for Supervised Fine-Tuning (SFT) to "cold start" the model's performance.
\end{itemize}
\textbf{Candidate Task Mining}. The candidate task pool is primarily derived from real-world GitHub issues, Pull Requests (PRs), and commit metadata. To ensure the data is suitable for model training, we apply a series of heuristic filters—such as the clarity of PR descriptions and the number of files modified—to select tasks with an appropriate level of complexity. Notably, our tasks are not restricted to Python. Instead, our pipeline encompasses 12 mainstream programming languages. To guarantee the quality and maturity of the source repositories, we enforce strict selection criteria: each repository must have over 10 stars and a history of at least two successfully merged Pull Requests. The tasks are categorized into several distinct engineering domains to ensure diverse functional coverage:
\begin{itemize}
\item \textbf{Bug Fix} Identifying, diagnosing, and resolving defects within the source code.
\item \textbf{Feature Enhancement} Improving or expanding existing functionality to add value to the project.
\item \textbf{Refactoring} Modifying source code to optimize internal structure without changing external behavior.
\item \textbf{Test Case Generation} Automatically generating unit tests and utilizing test suites to verify code integrity.
\end{itemize}
\textbf{Verifiable Environment Construction}. To support the training requirements of SWE tasks, we have developed a large-scale infrastructure for execution sandboxes. In practice, we observed that simply constructing a Docker image capable of hosting the repository is insufficient; the execution of test suites often fails due to unresolved package dependencies and other environmental inconsistencies. Consequently, a repository-level image alone cannot adequately support the training workflow. To address this, we have decoupled the environment construction process into two distinct phases. One is the Docker Image Provisioning stage, which focuses on the baseline containerization of the repository. The other is the Test Case Validation stage, where we execute the test suites to verify whether the constructed environment is "verifiable" — ensuring that the dependencies are correctly configured and the environment is functional for downstream training and evaluation.
\begin{itemize}
\item \textbf{Phase 1: Build \& Initialize}. Manually constructing executable sandbox environments is both time-consuming and labor-intensive. To address this, we employ an Agent-based approach, leveraging autonomous agents to build Docker images directly from existing GitHub repositories. The construction process utilizes an agentic automation workflow: the agent iteratively attempts to build the image, while human intervention is introduced only for execution failures. Through this process, we extract and refine successful patterns into a reusable Skill Library. By continuously repeating this agentic loop, the system's efficiency and success rate improve over time. Upon the completion of an image build, a mandatory "smoke test" is executed. This serves as a preliminary validation step to determine whether the environment possesses the foundational operational capabilities required for subsequent tasks. 
\item \textbf{Phase 2: Test Execution \& Verification}. We employ an Agent-based approach to locate relevant test cases and generate executable test commands. By testing the project code in its states both before and after the Pull Request (PR), we identify Pass-to-Pass (P2P) and Pass-to-Fail (P2F) transitions, which serve as the primary success/failure signals. To provide more precise reward signals for a given patch, we developed a multilingual test log parser. This tool is utilized not only to extract test results but also to evaluate the test coverage of the aforementioned schemes. Based on these metrics, we systematically prune Docker images with insufficient coverage to ensure the quality.
\end{itemize}
\textbf{Building Trajectories}. To maximize the utility of the verifiable environments, we utilized the SWE-Smith \cite{yang2025swesmith} framework to synthesize a batch of tasks. To ensure sufficient task complexity, we filtered out overly simplistic cases by monitoring the number of modified lines and edited files. Moving forward, the pipeline bifurcates into two distinct branches:
\begin{itemize}
\item \textbf{Trajectory Rollout}. We utilize multiple agent frameworks (e.g., OpenHands \cite{openhands}, SWE-agent \cite{yang2024sweagent}, mini-swe-agent \cite{yang2024sweagent}) to generate (rollout) interaction trajectories that successfully pass the predefined test cases. 
\item \textbf{Atomic Capability Task Generation}. We derive specialized tasks focused on the atomic capabilities of an agent, such as precise code editing and automated test case generation.
\end{itemize}

\subsubsection{Tool-Integrated Reasoning}


Tool-Integrated Reasoning (TIR) enhances Large Language Models (LLMs) by incorporating external tool use into their reasoning process. Unlike traditional models that rely exclusively on pre-trained parametric knowledge, TIR enables models to decide when to invoke tools and to generate specific instructions—such as Python code or search queries. The execution results are then fed back into the model to inform subsequent reasoning and response generation. This iterative approach improves computational accuracy and information freshness, effectively addressing the inherent limitations of LLMs like calculation errors and knowledge cutoffs.

In this section, we construct and analyze four specialized TIR datasets tailored to distinct functional domains. We develop an automated, scalable data synthesis pipeline to generate high-quality reasoning trajectories. This pipeline focuses on capturing the specific reasoning patterns needed for Code Interpretation, Agentic Search, and hybrid scenarios that require the coordinated use of both tools.
Although our proposed JoyAI-LLM Flash is an instruct model, we find adding reasoning/thinking data in the SFT stage can improve the non-thinking capacity of the instruct model. 


\paragraph{Code-Centric Trajectories.} We extract complex mathematical problems and, where available, their corresponding ground truths from datasets such as OpenR1-Math-220k \cite{openr1_math_220k} and Nemotron-Math-v2 \cite{du2025nemotron}. Using DeepSeek-V3.2 \cite{deepseekai2025deepseekv32}, we distill these into TIR trajectories.

We have developed an interactive environment between the model and the Python interpreter, allowing the LLM to dynamically utilize the Python interpreter for iterative and symbolic computations. Our Python setup includes essential libraries that enable robust mathematical operations and solving capabilities. For example, the setup employs the \texttt{math} library for basic arithmetic and \texttt{sympy} for symbolic mathematics, with commands demonstrating tasks like initializing symbols and solving equations using \texttt{from sympy import symbols, Eq, solve}.

During data distillation, the model is restricted to a maximum of 20 tool invocations per problem session. Occasionally, the model may produce erroneous code, which the interpreter catches and returns as tool responses, aiding the model in debugging and retrying. After distillation, incomplete trajectories, as well as those containing plotting code such as \texttt{matplotlib}, are rigorously filtered out.

This process culminates in a comprehensive dataset of multi-round TIR records, integrating Python tools and providing a substantial foundation to assess and enhance LLM tool-integrated reasoning capabilities.

\paragraph{Search-Centric Trajectories.} To cultivate robust information-seeking, multi-hop reasoning, and agentic tool-use capabilities, we construct a search-centric trajectory corpus by distilling execution traces from DeepSeek-V3.2 \cite{deepseekai2025deepseekv32} across three data sources: \texttt{Complex QA \& Agentic Benchmarks} (~6,601 queries aggregated from seven established benchmarks including 2WikiMultihopQA \cite{2wiki}, MuSiQue \cite{musique}, Bamboogle \cite{bamboolge}, SimpleQA \cite{simpleqa}, FRAMES \cite{frames}, ScholarSearch \cite{scholarsearch}, and GAIA \cite{gaia}, covering multi-hop reasoning, factuality, complex RAG, and real-world agentic tasks), \texttt{TaskCraft} \cite{taskcraft} (17K search-relevant instances selected from a large pool of tool-intensive agentic tasks spanning single-step to expert-level multi-step executions), and \texttt{Nemotron-Science-v1} \cite{NemotronPostTrainingDatasetV1} (20K sampled instances from a multiple-choice scientific reasoning corpus). For sources with verifiable ground-truth labels, we retain only trajectories with correct final answers. The resulting corpus comprises trajectories with an average of 8.64 search invocations each.



Rather than exposing the model to raw search results, we introduce a summary agent as the sole interface to search results. Upon each search invocation, the model issues a structured call with explicit search keywords and an intent statement; the summary agent returns a concise, query-focused synthesis of the retrieved web pages. This design prevents overly long contexts, reducing computational overhead while preserving model performance during both training and inference.

\paragraph{Hybrid Tool-Integrated Trajectories.} Beyond task-specific datasets, we further curate a sophisticated category of trajectories that necessitate the synergistic coordination of both code interpreters and search engines. In these complex scenarios, the model must exhibit high-level planning: typically utilizing Agentic Search to retrieve specialized domain knowledge or external constants, followed by Code Interpretation to perform rigorous algorithmic verification or numerical modeling based on the retrieved data.

\paragraph{Terminal-Centric Trajectories.} To enhance the model’s proficiency in terminal-based operations, we synthesize a diverse set of task scenarios within a standardized, constrained Docker environment. Despite the restricted scope of the environment, we achieve high task diversity by utilizing capability decomposition and evolutionary sampling to expand from initial seed data. We employ DeepSeek-V3.2 \cite{deepseekai2025deepseekv32} to generate reasoning-action trajectories for each task. To ensure data quality, we implement an automated validation pipeline in which an LLM-based judge evaluates each trajectory across five key dimensions: completion, correctness, efficiency, safety, and overall quality. Trajectories with low scores are strictly excluded. This filtering mechanism ensures that only functionally viable and safe data remain for training, providing a stable yet diverse signal for subsequent SFT.


By constructing these four TIR datasets, we aim to comprehensively evaluate and improve models' abilities to integrate diverse external tools into their reasoning process, thereby pushing the frontier of automated, tool-augmented artificial intelligence.

\subsection{Direct Preference Optimization (DPO)}
During the Direct Preference Optimization (DPO) phase, we train the model for 1,000 steps utilizing a learning rate of 1e-6 with cosine-decay to 1e-7 and a global batch size of 256. To construct the DPO dataset, we curate STEM, general conversational, and safety queries distinct from those of the SFT stage. We form preference pairs by contrasting high-quality responses with negative examples derived from rejection sampling during SFT, specifically targeting prevalent failure cases such as hallucinations and instruction-following deviations. Ultimately, the DPO stage is crucial to the model's final performance; its rapid convergence provides a highly efficient mechanism to systematically penalize and eliminate undesirable outputs prior to RL.

\subsection{Reinforcement Learning (RL)}
\label{subsec:rl}

\input{fiberpo_main}

\input{fiberpo_exp}

\subsection{Instruct Model Evaluation}

To provide a comprehensive assessment of JoyAI-LLM Flash, we evaluate the model on a diverse set of widely used LLM benchmarks covering multiple capabilities, including
\begin{itemize}
    \item \textbf{General Knowledge}: MMLU~\cite{hendrycks2020measuring}, MMLU-Pro~\cite{wang2024mmlu}, HellaSwag~\cite{zellers2019hellaswag}, CMMLU~\cite{li2024cmmlu}, C-Eval~\cite{huang2023c}, GPQA-Diamond~\cite{rein2024gpqa}, SuperGPQA~\cite{pteam2025supergpqa}.

    \item \textbf{Math Reasoning}: MATH-500~\cite{hendrycks2021measuring}, AIME$^\prime$25.

    \item \textbf{Coding Ability}: HumanEval~\cite{chen2021evaluating}, LiveCodeBench v6~\cite{jain2024livecodebench}, SWE-bench Verified~\cite{jimenez2024swebench}.

    \item \textbf{Instruction Following}: AlignBench~\cite{liu-etal-2024-alignbench}, IFEval~\cite{zhou2023ifeval}.

    \item \textbf{Long-context Ability}:  RULER~\cite{hsieh2024ruler}.

    \item \textbf{General Tasks}:
    LiveBench$_\text{2024-11-25}$~\cite{livebench}.

    \item \textbf{Agent \& OpenClaw}:
    $\tau^2$-Bench~\cite{barres2025tau2}, PinchBench.
\end{itemize}

The experimental results are shown in Table~\ref{tab:instruct_eval}, where Qwen3-Next-80B-A3B, Qwen3.5-35B-A3B, and Qwen3-30B-A3B are the baseline instruct models. We also include GLM-4.7-Flash-Thinking for comparison, as its reasoning capabilities provide a more direct alignment with our JoyAI-LLM Flash than its standard instruct counterpart.

\begin{figure*}[htb]
    \centering
    \begin{subfigure}[b]{0.33\linewidth}
        \includegraphics[trim=0pt 0pt 0pt 0pt,clip,width=\textwidth]{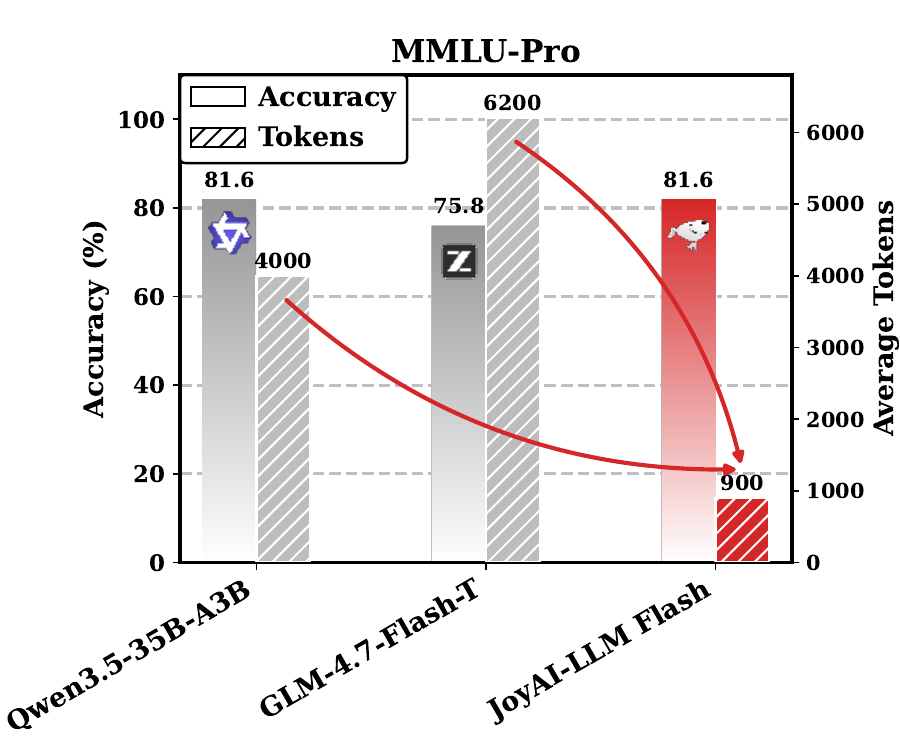}
    \caption{}
    \end{subfigure}\hfill
    \begin{subfigure}[b]{0.33\linewidth}
        \includegraphics[trim=0pt 0pt 0pt 0pt,clip,width=\textwidth]{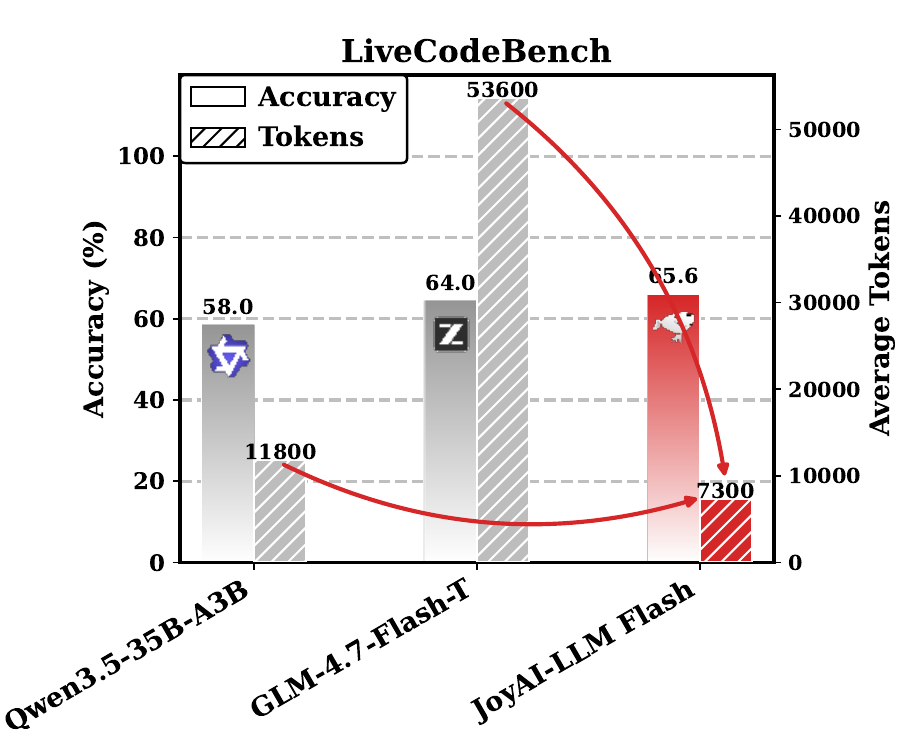}
    \caption{}
    \end{subfigure}\hfill
    \begin{subfigure}[b]{0.33\linewidth}
        \includegraphics[trim=0pt 0pt 0pt 0pt,clip,width=\textwidth]{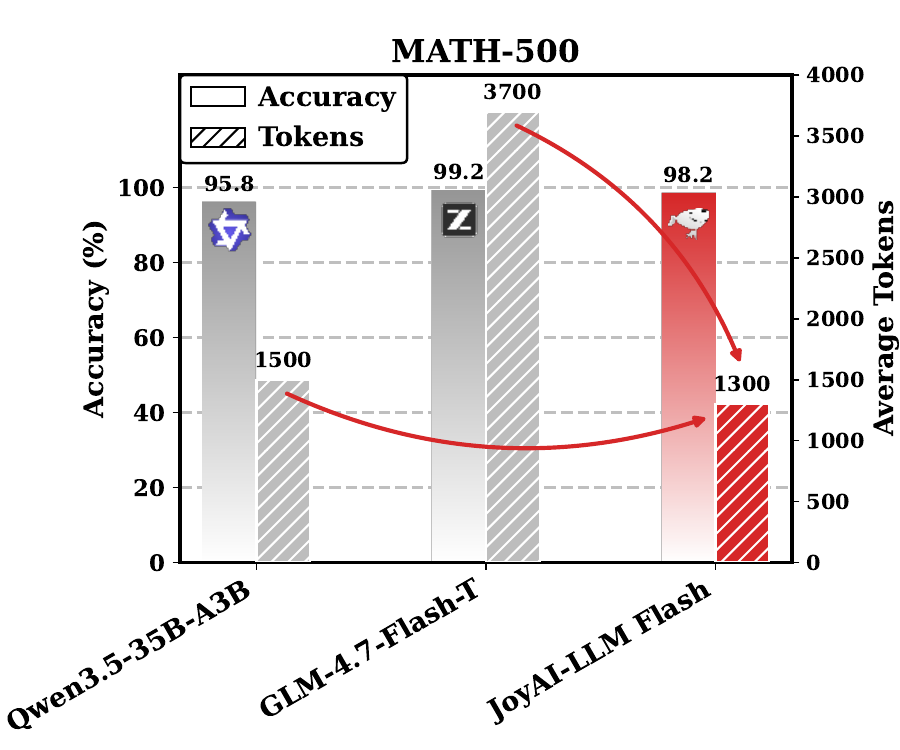}
    \caption{}
    \end{subfigure}\\
    \begin{subfigure}[b]{0.33\linewidth}
        \includegraphics[trim=0pt 0pt 0pt 0pt,clip,width=\textwidth]{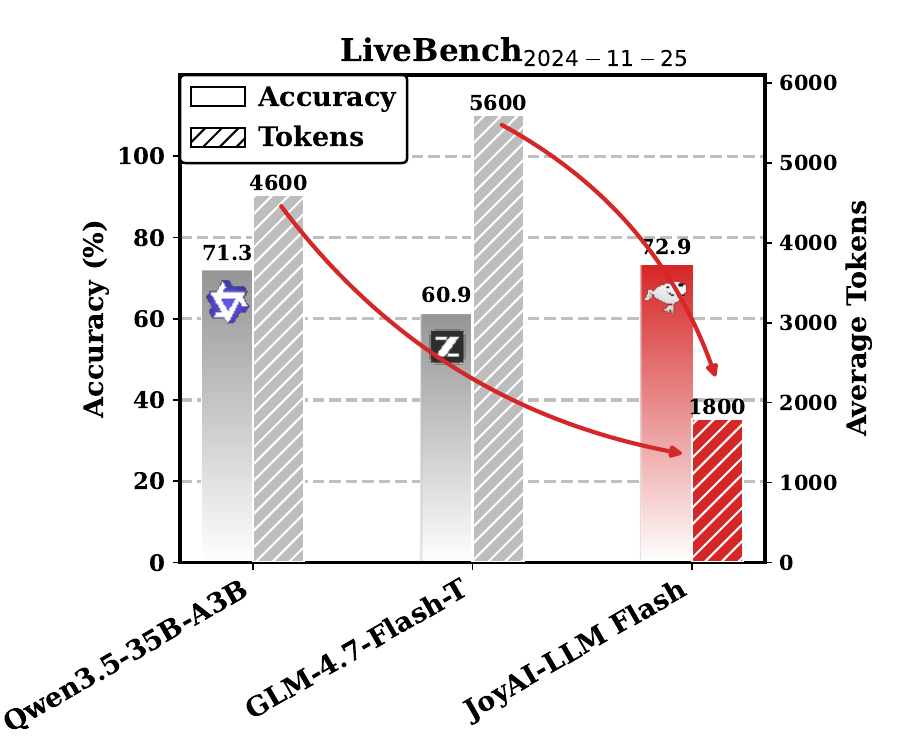}
    \caption{}
    \end{subfigure}\hfill
    \begin{subfigure}[b]{0.33\linewidth}
        \includegraphics[trim=0pt 0pt 0pt 0pt,clip,width=\textwidth]{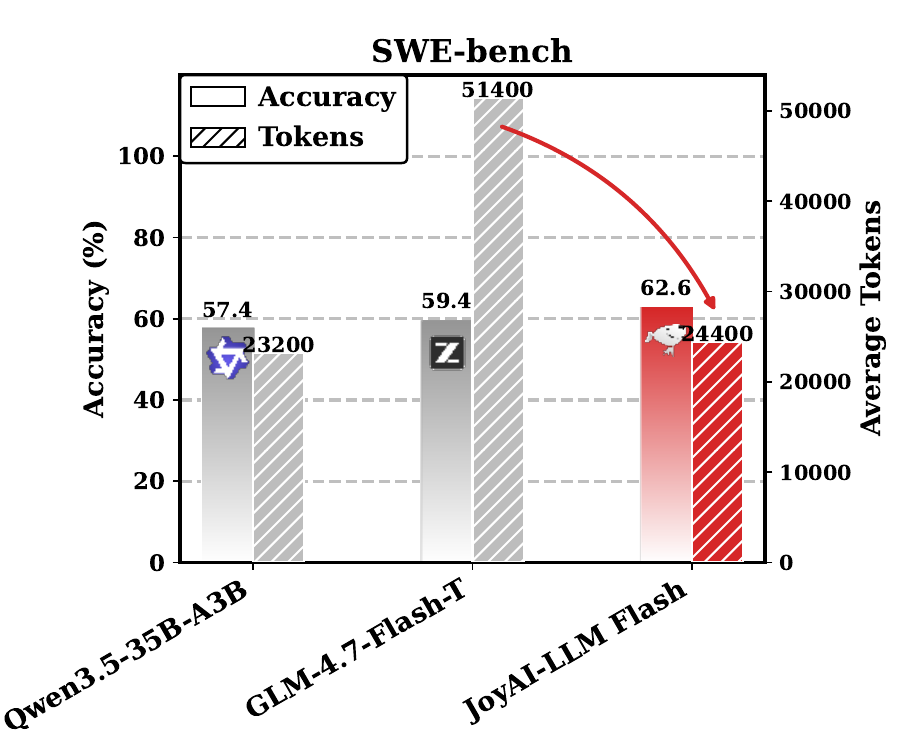}
    \caption{}
    \end{subfigure}\hfill
    \begin{subfigure}[b]{0.33\linewidth}
        \includegraphics[trim=0pt 0pt 0pt 0pt,clip,width=\textwidth]{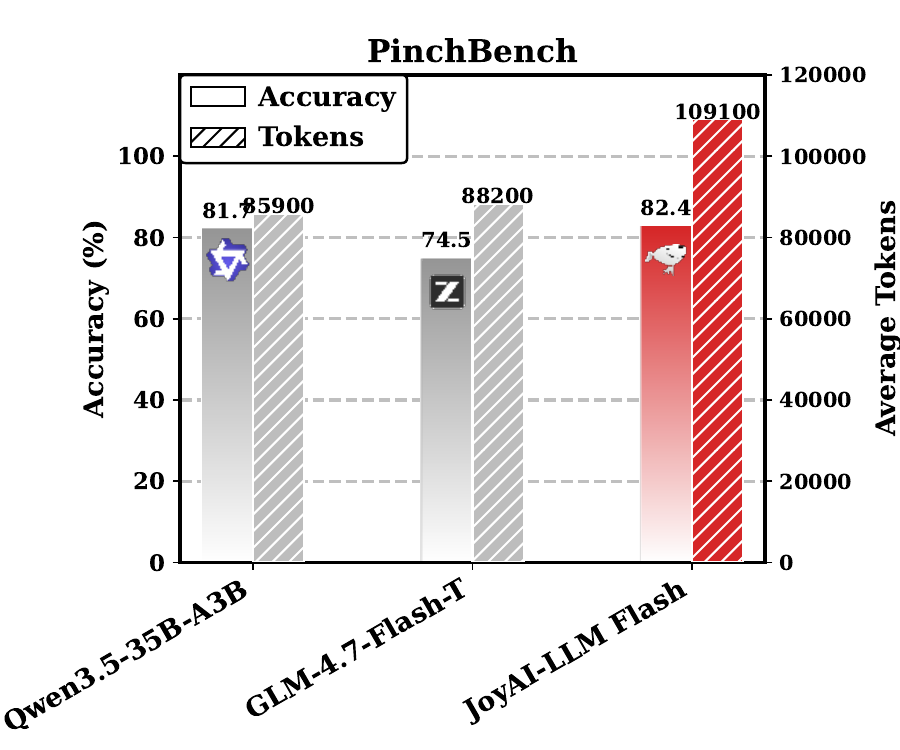}
    \caption{}
    \end{subfigure}
\caption{Comparison of model performance (left bars) and token consumption (right bars) across six benchmarks. Qwen3.5-35B-A3B and JoyAI-LLM Flash are instruct models, ``GLM-4.7-Flash-T'' refers to GLM-4.7-Flash-Thinking, which is included due to its comparable performance.}
\label{fig:acc_vs_token_bar}
\end{figure*}

As can be observed, our JoyAI-LLM Flash achieves remarkable token efficiency. Specifically, on the LiveCodeBench, JoyAI-LLM Flash surpasses GLM-4.7-Flash-Thinking by 1.6\% accuracy with a 85\% reduction in token usage. Beyond efficiency, JoyAI-LLM Flash also demonstrates strong competitive performance across diverse tasks, including mathematics and long-context understanding. 
The token efficiency of JoyAI-LLM Flash is illustrated in Figure \ref{fig:acc_vs_token_bar}. Specifically, the accuracy (left bar) and token usage (right bar) are compared across different models, highlighting our efficiency gains.
Notably, although JoyAI-LLM Flash consumes more tokens on PinchBench, it achieves the best accuracy compared to other models.

\begin{table}[htbp]
    \centering
    \definecolor{graybg}{gray}{0.9} 
    \small 
    \setlength{\tabcolsep}{3.5pt} 
    \caption{Comparison with baseline models. Qwen3-Next-80B-A3B, Qwen3.5-35B-A3B, and Qwen3-30B-A3B are instruct models; ``GLM-4.7-Flash-T'' refers to GLM-4.7-Flash-Thinking, which is additionally included due to its more comparable performance. Results marked with asterisk$^*$ are directly cited from original papers and differ substantially from our reproduced results.}
    \begin{tabular}{lcccccccccc}
        \toprule
        \multirow{2}{*}{\textbf{Task}} & \multicolumn{2}{c}{\rotatehead{Qwen3-Next-80B-A3B}} & \multicolumn{2}{c}{\rotatehead{Qwen3.5-35B-A3B}} & \multicolumn{2}{c}{\rotatehead{Qwen3-30B-A3B}} & \multicolumn{2}{c}{\rotatehead{GLM-4.7-Flash-T}} & \multicolumn{2}{c}{\rotatehead{JoyAI-LLM Flash}} \\
         & Acc & \#Token & Acc & \#Token & Acc & \#Token & Acc & \#Token & Acc & \#Token \\
        \midrule

        \rowcolor{graybg}
        \multicolumn{11}{l}{\textbf{General Knowledge}} \\ 
        \quad MMLU       & 86.4 & 400 & \textbf{91.2} & 900 & 88.0 & \textbf{200} & 88.2 & 2000 & 89.1 & \textbf{200}  \\
        \quad MMLU-Pro   & 76.7 & 1500 & \textbf{81.6} & 4000 & 78.7 & 1600 & 75.8 & 6200 & \textbf{81.6} & \textbf{900}   \\
        \quad HellaSwag  & 88.1 & \textbf{<100} & 89.1 & \textbf{<100} & 86.2 & \textbf{<100} & 71.5 & 3000 & \textbf{91.7} & \textbf{<100} \\
        \quad CMMLU      & 89.1 & 500 & \textbf{89.5} & 600 & 87.1 & 500 & 80.2 & 4900 & 86.7 & \textbf{400}  \\
        \quad C-Eval     & 89.2 & 700 & \textbf{91.5} & 800 & 87.8 & 800 & 77.0 & 12200 & 88.7 & \textbf{600} \\
        \quad GPQA-Diamond  & 73.0 & 4000 & 74.2 & 4300 & 66.2 & 4100 & \textbf{76.7} & 19900 & 74.5 & \textbf{2800} \\
        \quad SuperGPQA  & 60.0 & 1900 & \textbf{62.0} & 3700 & 53.0 & 2000 & 41.0 & 8400 & 55.7 & \textbf{1300} \\
        \midrule

        \rowcolor{graybg}
        \multicolumn{11}{l}{\textbf{Math}} \\
        \quad MATH-500   & 97.4 & 1700 & 95.8 & 1500 & 97.6 & 1400 & \textbf{99.2} & 3700 & 98.2 & \textbf{1300}  \\
        \quad AIME$^\prime$25   & 70.4 & 6900 & 56.7 & 7000 & 63.8 & 6100 & \textbf{92.1} & 26000 & 72.9 & \textbf{5400} \\
        \midrule

        \rowcolor{graybg}
        \multicolumn{11}{l}{\textbf{Coding}} \\
        \quad HumanEval  & \textbf{95.1} & 600 & 93.9 & \textbf{300} & 92.1 & 400 & 93.9 & 7200 & 94.5 & 900  \\
        \quad LiveCodeBench v6& 58.8 & 8500 & 58.0 & 11800 & 48.1 & 15900 & 64.0$^*$ & 53600 & \textbf{65.6} & \textbf{7300}  \\
        \quad SWE-bench Verified & 31.2 & 25400 & 57.4 & 23200 & 29.0 & \textbf{16400} & 59.4$^*$ & 51400 & \textbf{62.6} & 24400 \\
        \midrule

        \rowcolor{graybg}
        \multicolumn{11}{l}{\textbf{Instruction Following}} \\
        \quad AlignBench & \textbf{8.3} & 800 & 8.1 & 1000 & 7.9 & 1200 & 6.9 & 3600 & 8.0 & \textbf{700} \\
        \quad IFEval     & 84.7 & 600 & 81.8 & 1000 & 80.8 & 500 & \textbf{85.4} & 1900 & 82.8 & \textbf{400} \\
        \midrule

        \rowcolor{graybg}
        \multicolumn{11}{l}{\textbf{Long-Context}} \\
        \quad RULER      & 94.2 & 100 & \textbf{96.0} & \textbf{<100} & 93.7 & \textbf{<100} & 74.7 & 8300 & 95.7 & \textbf{<100}  \\
        \midrule
        \rowcolor{graybg}
        \multicolumn{11}{l}{\textbf{General Tasks}} \\
        \quad LiveBench$_\text{2024-11-25}$  & \textbf{75.9} & 2300 & 71.3 & 4600 & 68.5 & 2700 & 60.9 & 5600 & 72.9 & \textbf{1800} \\
        \midrule
        \rowcolor{graybg}
        \multicolumn{11}{l}{\textbf{Agent \& OpenClaw}} \\
        \quad $\tau^2$-Bench & 38.4 & 2200 & \textbf{76.6} & 2800 & 31.6 & \textbf{1900} & 69.9 & 3400 & 74.1 & 3000 \\
        \quad PinchBench & \textbf{83.7} & 107700 & 81.7 & \textbf{85900} & 67.0 & 210400 & 77.1 & 145000 & 82.4 & 109100 \\
        \bottomrule
    \end{tabular}
    \vspace{10pt}
    \label{tab:instruct_eval}
\end{table}

\clearpage
\section{Inference}
\label{sec:inference}
The architecture of JoyAI-LLM Flash deliberately adopts a compact parameter scale of 48 billion together with a highly sparse Mixture-of-Experts (MoE) structure.
Meanwhile, we employ a co-design of training and inference optimizations, including Quantization-Aware Training (QAT) and dense Multi-Token Prediction (MTP).
Furthermore, we evaluate the inference throughput performance across various context lengths under the prefill–decode disaggregation setting.

\subsection{Quantization}
JoyAI-LLM Flash adopts both Quantization-Aware Training (QAT) and Post-Training Quantization (PTQ) to achieve the optimal trade-off between model accuracy and throughput.
All quantized models are open-sourced on \href{https://huggingface.co/collections/jdopensource/JoyAI-LLM Flash}{HuggingFace}.

\begin{figure}[b]
  \centering
  \includegraphics[width=0.75\linewidth]{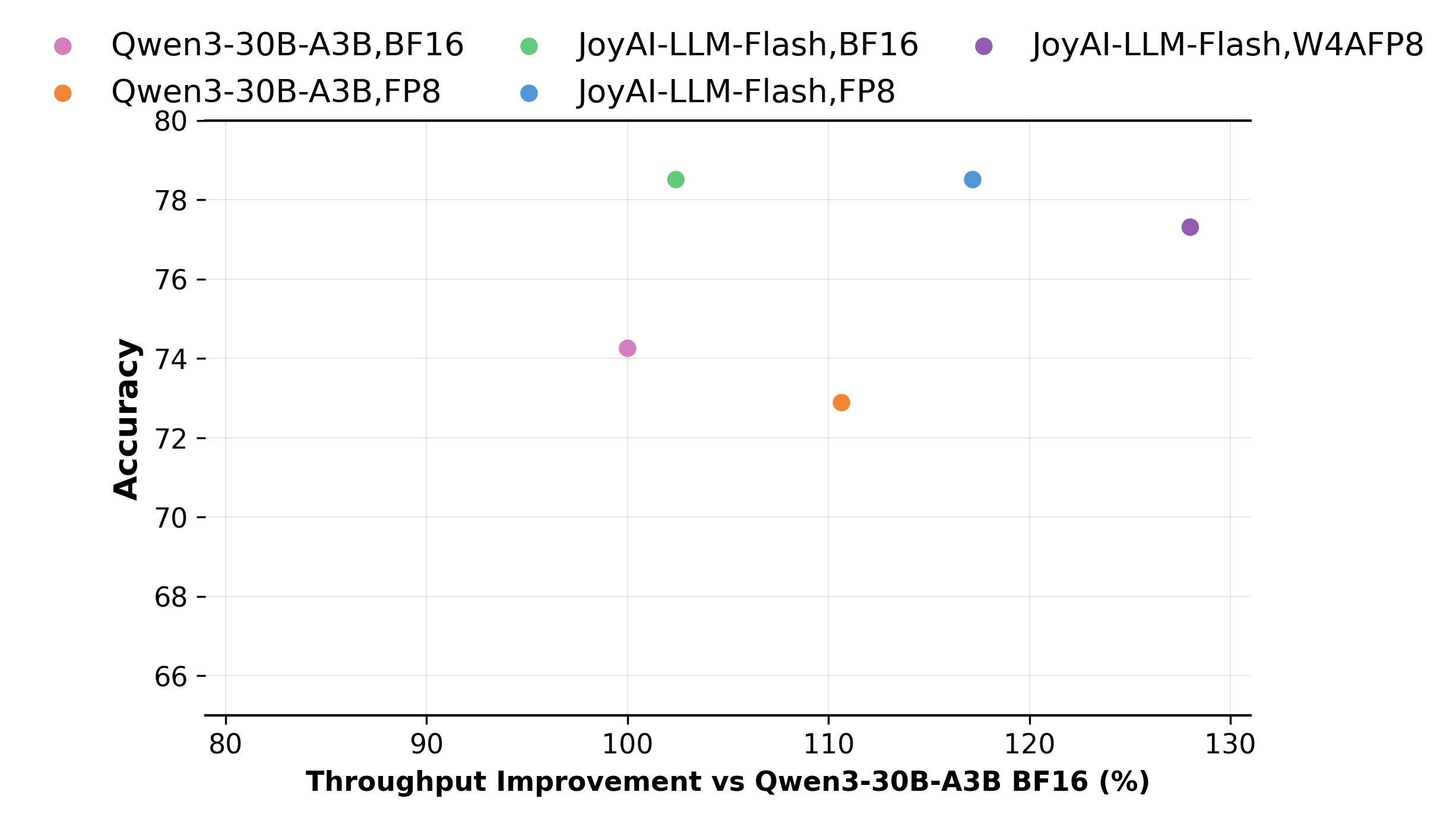}
  \caption{Comparison of Accuracy and Throughput for Quantized Models: JoyAI-LLM Flash vs. Qwen3-30B-A3B. The accuracy is measured by the mean accuracy across three distinct domain datasets: MATH-500, GPQA, and MBPP.}
  \label{fig:quant_performance_vs_qwen}
\end{figure}

During the QAT phase, JoyAI-LLM Flash simulates INT4 quantization during training by inserting fake-quantization operators (Quantize $\rightarrow$ DeQuantize $\rightarrow$ Quantize) and keeping weights in high precision for stable optimization. 
To cope with non-differentiable operations such as rounding and clamping, we adopt Straight-Through Estimators (STE) \cite{yin2019understanding} in backpropagation, allowing gradients to flow to the master weights. 
This aligns with the practice of mainstream LLMs such as Kimi-K2-Thinking~\cite{kimi_k2_thinking} and GLM-5~\cite{glm5_arxiv}.
A beneficial side effect is the stabilization of the reinforcement learning stage: INT4 quantization fosters more robust model rollouts and reduces noise diversity by narrowing the numerical space.
JoyAI-LLM Flash thus maintains high accuracy even when applying the simple Round-To-Nearest quantization scheme at lower bit widths.
During the PTQ phase, 
we compare JoyAI-LLM Flash with Qwen3-30B-A3B BF16 baseline under BF16, FP8, and W4AFP8~\cite{modelopt_quant} quantization in Figure \ref{fig:quant_performance_vs_qwen}.
All the experiments were conducted using vLLM~\cite{vllm_repo} and TRT-LLM~\cite{tensorrtllm_repo}. We reported the best out of the two for each model.
Model accuracy is evaluated by the mean accuracy across three distinct domain datasets: MATH-500\cite{hendrycks2021measuring}, GPQA~\cite{rein2024gpqa}, and MBPP~\cite{austin2021program}.
The results indicate that JoyAI-LLM Flash architecture consistently outperforms the Qwen baseline in terms of accuracy and throughput.
Qwen3-30B-A3B FP8 model yields a throughput gain of 10\%, while suffering noticeable accuracy degradation.
In contrast, JoyAI-LLM Flash FP8 model improves throughput by 17\% with nearly no accuracy degradation. Moreover, the W4AFP8 model maximizes the throughput gain of nearly 28\% over the baseline with a slight accuracy drop of 1.2\%. 
These findings demonstrate that JoyAI-LLM Flash achieves an optimal trade-off between accuracy and efficiency, even with 1.63× larger model weights than the Qwen baseline.

Furthermore, to accommodate model usage on edge devices, we also released the effective low-bit GGUF~\cite{ggufformat} variants of JoyAI-LLM Flash.
Inspired by the practices of NVIDIA NVFP4 quantization and GGUF’s K-Quants, we propose a “DoubleQuant” strategy tailored for less sensitive weights: 
partition the weight matrix into super blocks, 
apply the aforementioned quantization method within each block, 
perform a global-like quantization across blocks and store the double-quantized scales at lower precision (e.g., 6-bit or 8-bit).
Experimental results show that this DoubleQuant strategy achieves comparable accuracy to the BF16 baseline, validating the effectiveness of the proposed approach.

\subsection{Multi-Token Prediction}
JoyAI-LLM Flash features a lightweight dense MTP architecture, achieving state-of-the-art speedup despite a medium acceptance length.
Table \ref{tab:specbench_mtp3_speedup} evaluates the MTP performance of JoyAI-LLM Flash on SpechBench~\cite{xia-etal-2024-unlocking} under the MTP 3 layers and concurrency 64 configuration. The performance is evaluated by the acceptance length, ratio and speedup over the non-MTP counterpart. 
We compare JoyAI-LLM Flash with a suite of MTP-optimized LLMs, including
Qwen3.5-35B-A3B~\cite{qwen3.5}, 
Step-3.5-Flash~\cite{huang2026step35flashopen},
MiMo-V2-Flash~\cite{xiao2026mimo},
GLM-5~\cite{glm5team2026glm5vibecodingagentic},
GLM-4.7-Flash~\cite{5team2025glm45agenticreasoningcoding},
DeepSeek-V3.2~\cite{deepseekai2025deepseekv32},
and DeepSeek-V3~\cite{liu2024deepseek}. 
JoyAI-LLM Flash achieves the highest speedup of 1.87×, representing a 3\% improvement over the closest competitor, GLM-5 (1.82×), and a 72\% improvement over the slowest model, GLM-4.7-Flash (1.09×).

\begin{table}[htbp]
  \centering
  \caption{SpecBench MTP-3 Speculative Decoding Performance. Best results are marked in
bold.}
  \label{tab:specbench_mtp3_speedup}
  \begin{tabular}{c c c c c c c c}
    \toprule
    \multirow{2}{*}{\textbf{Model}} & 
    \multirow{2}{*}{\textbf{GPU}} & 
    \multirow{2}{*}{\textbf{Speedup}} &
    \multicolumn{2}{c}{\textbf{Acceptance}} & \multicolumn{2}{c}{\textbf{TPS}} \\
    \cmidrule(lr){4-5} \cmidrule(lr){6-7}
    & & & \textbf{length} & \textbf{ratio} & \textbf{user} & \textbf{server} \\
    \midrule
    \multicolumn{1}{c|}{JoyAI-LLM Flash}  & \multicolumn{1}{c|}{1} & \textbf{1.87×} & 2.20 & 40.35 & 66 & 4241 \\
    \multicolumn{1}{c|}{GLM-5~\cite{glm5team2026glm5vibecodingagentic}}            & \multicolumn{1}{c|}{8} & 1.82× & 3.03 & \textbf{75.84} & 31 & 1969 \\
    \multicolumn{1}{c|}{DeepSeek-V3.2~\cite{deepseekai2025deepseekv32}}    & \multicolumn{1}{c|}{8} & 1.79× & 2.55 & 63.72 & 31 & 1958 \\
    \multicolumn{1}{c|}{Qwen3.5-35B-A3B~\cite{qwen3.5}}    & \multicolumn{1}{c|}{1} & 1.61× & \textbf{3.18} & 72.56 & \textbf{85} & \textbf{5428} \\
    \multicolumn{1}{c|}{MiMo-V2-Flash~\cite{xiao2026mimo}}    & \multicolumn{1}{c|}{4} & 1.61× & 2.69 & 67.22 & 47 & 3033 \\
    \multicolumn{1}{c|}{Step-3.5-Flash~\cite{huang2026step35flashopen}}   & \multicolumn{1}{c|}{4} & 1.39× & 2.21 & 55.23 & 48 & 3048 \\
    \multicolumn{1}{c|}{DeepSeek-V3~\cite{liu2024deepseek}} & \multicolumn{1}{c|}{8} & 1.21× & 2.71 & 56.86 & 25 & 1608 \\
    \multicolumn{1}{c|}{GLM-4.7-Flash~\cite{5team2025glm45agenticreasoningcoding}}    & \multicolumn{1}{c|}{1} & 1.09× & 2.11 & 36.84 & 26 & 1695 \\
    \bottomrule
  \end{tabular}
\end{table}

\begin{figure}[b]
  \centering
  \includegraphics[width=0.75\linewidth]{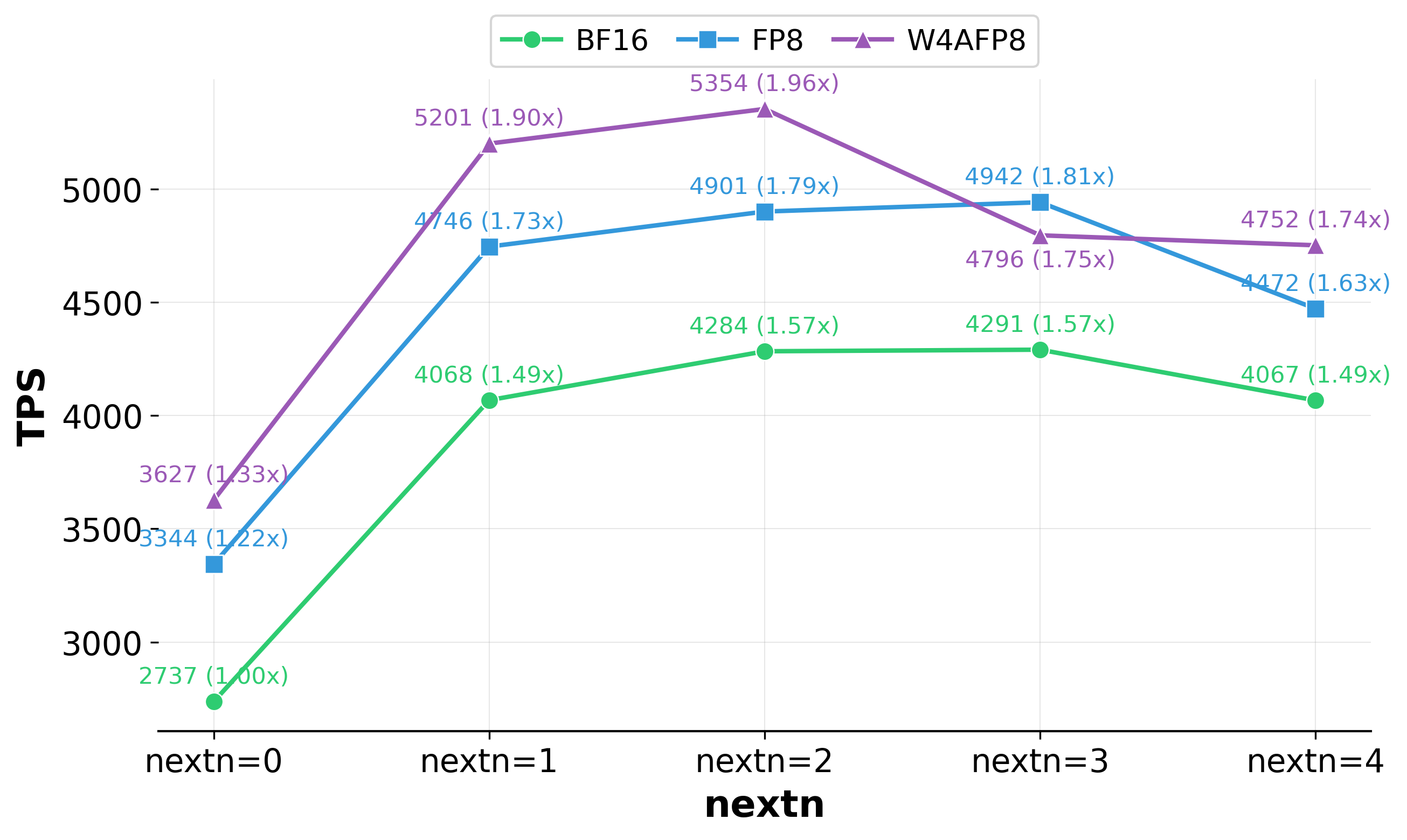}
  \caption{Joint Optimization of MTP and Quantization (ISL/OSL = 1K/2K, Concurrency = 64)}
  \label{fig:mtp_performance}
\end{figure}

\label{sec:mpt_quant}
To better align with user habits, we integrate MTP with quantization and conduct joint optimization.
The overall performance of MTP is evaluated by the \textit{next-n} configuration. 
We evaluate inference throughput across three quantization formats—BF16, FP8, and W4AFP8—on a randomly sampled dataset with ISL=1K and OSL=2K.
All quantization settings were deployed with TensorRT-LLM~\cite{tensorrtllm_repo}, integrated with CUDA Graph, Torch Compile, and Overlap Scheduling techniques.
As shown in Figure \ref{fig:mtp_performance}, both BF16 and FP8 model achieved optimal throughput under the MTP setting of \textit{next-n}=3. 
Compared to the baseline configuration (BF16 with \textit{next-n}=0), these settings delivered a throughput acceleration of 1.57× and 1.81×, respectively. 
The overall best performance was observed with W4AFP8 quantization at \textit{next-n}=2, yielding a 1.96× speedup over the same baseline. 
The experiments indicate that excessively high MTP layers (e.g., \textit{next-n}>3) could introduce diminishing or even negative returns, which is attributed to the additional computational overhead incurred in predicting multiple future tokens.

\subsection{Serving and Scheduling}

Modern LLM inference workloads exhibit highly variable sequence lengths, bursty arrival patterns, and drastically distinct compute–memory profiles between prefill and decode stages. 
And the serving efficiency is usually workload-dependent: short-context requests are governed by the trade-off of Time To First Token (TTFT) and Time Per Output Token (TPOT), whereas long-context requests are prefill-dominated and benefit more from cross-request prefix reuse. 
We therefore evaluate JoyAI-LLM Flash under these two distinct regimes.

Short-context workloads such as interactive chat typically prioritize user interaction experience and use TTFT and TPOT as the key performance metrics.
We build a workload grid with input lengths ranging from 128 to 2048 tokens and output lengths of 128 and 512 tokens. Request rates were varied from 1 to 8 requests per second (RPS).
JoyAI-LLM Flash maintains excellent responsiveness across all tested scales. 
To achieve comparable or better performance gains in real-world environments, we provide the following deployment insights based on our practical experience:

\begin{itemize}
\item \textbf{Intra-node PD Improves overall Performance:} For smaller models like JoyAI-LLM Flash, we recommend colocating prefill and decode instances on the same node to minimize communication overhead.
\item \textbf{Dynamic PD Adapts well to Real-time Workloads Variations:} Performance simulation tools such as AIConfigurator~\cite{aiconfigurator_repo} enable runtime dynamic scaling of PD instances according to SLA targets.
\end{itemize}

Long-context workloads such as Retrieval-Augmented Generation (RAG) and multi-turn agent tasks typically require large KV cache capacities.
However, limited memory frequently triggers KV cache eviction, which induces redundant recomputation and restricts cross-request KV reuse.
This constraint poses severe challenges for latency-sensitive metrics such as TTFT and degrades user experience.
To simulate this workload, we randomly generated data with no prefix reuse, featuring input lengths of 20,000 tokens and output lengths of 100 tokens. Request rates were varied from 0.25 to 3.0 requests per second (RPS). 
Experiments were conducted with a Prefill-Decode (PD) disaggregated deployment. And Mooncake~\cite{qin2024mooncake} was employed as a centralized KV cache store to manage cache across requests.
Also, we provide the following deployment insights under this scenario:
\begin{itemize}
\item \textbf{PD Disaggregation Delivers Better Flexibility:} 
Compared with aggregated deployment, PD disaggregation supports independent scaling of prefill and decode, and centralized KV caching lets us tune their instance ratio to better match workload demands.

\item \textbf{Choose KV Cache Management Wisely:} 
Two mainstream centralized KV cache management schemes exist: remote KV pooling and peer-to-peer (P2P) CPU sharing.
Selection depends on hardware infrastructure, with the goal of minimizing data transfer overhead.
Across both approaches, we strongly discourage TCP as the data-plane transport due to its high latency and overhead.

\item \textbf{Choose the Appropriate Cache Write Strategy:} 
The write strategy for KV cache should account for available bandwidth. 
When bandwidth is sufficient, we recommend writing to the distributed cache layer immediately upon each cache hit to maximize hit rates. 
While in scenarios with constrained I/O bandwidth, we recommend reducing write frequency and using an eviction policy to preserve hot data.

\item \textbf{Trade-off Between Recomputation and Transfer:} 
Smaller models exhibit higher sensitivity to data transfer overhead, even when KV caches are exchanged via P2P RDMA between instances.
Transfer latency can outweigh recomputation cost, making centralized KV cache management a net-negative optimization.
Careful evaluation is required to identify the crossover point for a given model and hardware setup.
\end{itemize}

\section{Conclusion and Future Work}
\label{sec:conclusion}
We present JoyAI-LLM Flash, a state-of-the-art medium-sized instruct language model with 3 billion activated parameters and 48 billion total parameters. JoyAI-LLM Flash was pretrained on 20 trillion text tokens using Muon optimizer, followed by large-scale supervised fine-tuning (SFT), direct preference optimization (DPO), and reinforcement learning (RL) across diverse environments. JoyAI-LLM Flash achieves strong performance across frontier knowledge, reasoning, coding tasks, and agentic capabilities. Moving forward, we aim to extend the model's paradigm by integrating continual learning and persistent memory, enabling the LLM to dynamically adapt and retain knowledge over time.

\newpage
\section{Contribution}
\paragraph{Project Leaders}
Chao Xue, Xiaodong He$^\dagger$\insert\footins{\noindent\footnotesize$^\dagger$Corresponding Author}

\paragraph{Core Contributors}
Bo Zhang, Bohua Cai, Chang Li, Chao Xue, Dongkai Liu, Guoqiang Huang, Jialong Shi, Liang Huang, Ming Ke, Panfeng Shi, Qi Wang, Qiaoqiao Yuan, Qiong Cao, Qixiang Wang, Rongcheng Bian, Shi Suo, Shijie Ren, Shijin Zhang, Shiying Fan, Shuai Xie, Tianyi Zhang, Wei Liu, Wentao Tan, Xiaodong He, Xuyang Peng, Ya Zhang, Yifei Liu, Yinhao Bai, Yuqi Zhang, Yuesong Zhang, Zhenfang Wang
 
\paragraph{Contributors}
Aichen Cai, Anmeng Zhang, Anson Li, Changjian Jiang, Changkai Lu, Chaocai Liang, Cheng Zhang, Fei Wang, Haijian Ke, Han Lin, Hao Wang, Ji Miao, Jiacheng Zhang, Jifeng Zhu, Jingjing Qian, Junhui Luo, Junwu Xiong, Lam So, Mingyang Li, Peng Hao, Qian Lai, Qingyu Yin, Rongduo Han, Shaoqiang Zheng, Shi Hu, Xianghan Meng, Xing Pan, Xiran Wang, Yanxu Chen, Yang Liu, Yangyang Duan, Yicheng Gong, Yidan Huang, Yongqiang Liu, Zerui Xie, Zhennan Shen, Zheyuan Liu, Zhuwei Zeng

\paragraph{Acknowledgment}
We thank Jiepeng Zhou, Kaiqing Lei, Shaoxiong Zhan, Tshihao Tsu, Yao Yao, Yaren Zhang, Yihui Wang, Zhengda Zhou, Zhenting Huang, and Zhihao Gong for their efforts. 

\appendix
\newpage

\bibliographystyle{unsrt}  
\bibliography{references}

\end{document}

%% file: fiberpo_main.tex
Large language models are no longer single, monolithic policies:
they are increasingly deployed and trained as heterogeneous
systems—agentic pipelines spanning domains and tools,
mixture-of-experts (MoE) architectures with conditional routing,
and distributed/asynchronous training stacks where optimization
noise and data nonstationarity are structural rather than
incidental. In this regime, alignment via
RLHF~\cite{ouyang2022training} must simultaneously handle
multi-scale instability: token-level stochasticity,
trajectory-level drift, and system-level heterogeneity
(domains/experts/agents) interacting in the same update. Existing
PPO-style ``proximal''
objectives~\cite{schulman2017proximal,shao2024deepseekmath,yu2025dapo}
provide only coarse local controls (mostly per-token clipping)
and limited diagnostics when failures arise from global structure
(e.g., a drifting subset of trajectories, an expert partition, or
a domain slice). This motivates importing more expressive
mathematical structure, beyond new loss heuristics, to build
controllers that can allocate stability budgets across the
relevant global contexts. In our JoyAI-LLM, we develop Fiber Bundle
Gating (FBG), a geometric framework grounded in fiber bundle
theory, and derive FiberPO from it, a concrete policy
optimization objective that decomposes trust-region maintenance
into compositional global and local components, providing
multi-scale stability control with first-order fidelity to the
true RL objective near on-policy and a restorative gradient
structure less explored in existing methods.

FiberPO rests on a principled theoretical foundation developed
in~\cite{fiberpo2026}. Classical TRPO~\cite{schulman2015trust}
trust regions collapse at the undiscounted horizon $\gamma{=}1$
required by LLM RL (the TRPO Vanishing Theorem
in~\cite{fiberpo2026}). This does not preclude trust-region-style
stabilization, but it shows that the classical radius cannot be used
as-is, necessitating a decoupling of how trust regions are maintained
(ratio clipping) from the specific radius prescribed by TRPO's
monotonic improvement guarantee. An intermediate result,
Aggregational Policy Censoring Objective (APC-Obj), achieves this
decoupling by proving that the clipping-based surrogate can exactly
reproduce trust-region updates (the APC-Obj-TRPO Equivalence Theorem
in~\cite{fiberpo2026}), so that the clipping mechanism remains
well-defined at any positive radius $\delta > 0$. APC-Obj also
provides a unified Ratio Gating Formalism from which
PPO~\cite{schulman2017proximal},
GRPO~\cite{shao2024deepseekmath}, and GSPO~\cite{zheng2025group}
are each derived as identified relaxations. This taxonomy reveals a
structural gap: token-wise methods (PPO, GRPO) do not bound
trajectory-level drift, while sequence-wise methods (GSPO) suppress
within-trajectory variation. To compose the two scales, we introduce
Fiber Bundle Gating (FBG), a geometric framework that organizes
tokens as a fiber bundle over trajectory-level contexts and
decomposes ratio gating into compositional base-level and fiber-level
operations, with provable first-order agreement with the true RL
objective near on-policy (the FBG First-Order Agreement Theorem
in~\cite{fiberpo2026}).

FiberPO is the concrete instantiation of FBG derived from the
APC-Obj objective through a sequence of controlled
transformations~\cite{fiberpo2026}. The FiberPO objective
factorizes each token's gated importance ratio into a
trajectory-level base weight and a token-level gated residual.
The base weight maintains a trust-region budget on
trajectory-level drift through a piecewise-linear aggregate gate
$g^{\rm agg}$, while the gated residual clips each token's
deviation from its trajectory mean via $\operatorname{logclip}$.
This two-scale decomposition provides independent control at both
levels, a structural property rarely explored in all prior
methods. Because fibrations compose algebraically, the same
gating mechanism extends to arbitrary hierarchical depth:
\cite{fiberpo2026} derives a Fibration Gating Hierarchy (FGH) and
instantiates FiberPO-Domain, a four-level variant with
independent trust-region budgets at the domain, prompt group,
trajectory, and token levels. In this report we present the
two-level trajectory-token case for conciseness. The full
theoretical development is given in~\cite{fiberpo2026}.


\subsubsection{The FiberPO Objective}
\label{subsubsec:fiberpo_obj}

Let $r_i := \pi_\theta(a_i|s_i)/\pi_{\theta_{\rm old}}(a_i|s_i)$ denote the importance ratio for token $i$, $\hat A_i$ the estimated advantage, and $T_\tau$ the length of trajectory $\tau$.
The augmented token space $\bar{\mathcal X} := \{(s_t(\tau),a_t(\tau),\tau,t)\}$ indexes each token by its trajectory membership and timestep.

\begin{definition}[FiberPO]\label{def:fiberpo_joyai}
The FiberPO objective is:
\begin{equation}\label{eq:fiberpo_obj}
    {\hat J^{{\text{FiberPO}}}}(\theta |\theta_{\rm old}) = \sum_{(s,a,\tau,t)\in\bar{\mathcal X}} {\frac{1}{{|\mathrm{Tj}^{\theta_{\rm old}}|}}{\frac{1}{{T_\tau}}}  \cdot\mathcal G(r_\bullet)_{s,a,\tau,t}\;\cdot \hat A_{s,a} },
\end{equation}
where $\mathrm{Tj}^{\theta_{\rm old}}$ is the set of sampled trajectories and the gating map $\mathcal G$ decomposes multiplicatively for each token $i \equiv (s,a,\tau,t)$:
\begin{equation}\label{eq:fiberpo_decomp}
    \mathcal G(r_\bullet)_i \;=\; \underbrace{\frac{\exp\circ\;g^{\rm agg}(\log s_\tau^+,\;C^+,\;T_\tau)}{\exp\circ\;g^{\rm agg}(\log s_\tau^-,\;C^-,\;T_\tau)}}_{w^{\rm base}_\tau\;:\;\text{base weight}} \;\cdot\; \underbrace{\frac{{\rm logclip}\!\left((s_\tau^{(l_i)})^{-l_i}\,r_i,\;\epsilon\right)}{{\rm logclip}\!\left((s_\tau^{(-l_i)})^{-l_i},\;\epsilon\right)}}_{\tilde r_i^{\rm fiber}\;:\;\text{gated residual}}.
\end{equation}
Each constituent is defined in detail below. At a high level, the
decomposition reflects the fiber bundle structure of sampled RLHF
data~\cite{fiberpo2026}: each token's log-ratio is split into a
trajectory-level component (how much the trajectory as a whole
has drifted) and a token-level residual (how much that token
deviates from its trajectory's drift). The base weight $w^{\rm
  base}_\tau$ corresponds to the base gate in the Fiber Bundle
Gating (FBG). It depends only on trajectory-level aggregate
log-ratios $s_\tau^+, s_\tau^-$, which separately track positive
and negative drift within each trajectory based on the sign label
$l_i := \operatorname{sign}(\log r_i) \in \{+1,-1\}$ (with
$s_\tau^{(l_i)}$ selecting the same-sign channel and
$s_\tau^{(-l_i)}$ the opposite), and is shared by all tokens in
trajectory $\tau$, controlling how much gradient signal the
trajectory as a whole is permitted to contribute through the
piecewise-linear gate $g^{\rm agg}$. The gated residual $\tilde
r_i^{\rm fiber}$ corresponds to the fiber gate. It captures each
token's deviation from the trajectory aggregate, gated by
$\operatorname{logclip}$ to prevent individual token spikes.
Together, the two components provide compositional multi-scale
control: the base weight maintains a trust-region budget at the
trajectory level, while the gated residual regulates per-token
outliers within each trajectory.
\end{definition}

\paragraph{Intuitive example.} To illustrate the practical
significance of this decomposition, consider two trajectories
answering ``Name a famous landmark'':
\begin{center}
\emph{``I love Paris and the Eiffel Tower''} \quad vs.\quad
\emph{``I love Rome and the Colosseum.''}
\end{center}
Globally (trajectory-level), the policy may strongly prefer the
Paris response, perhaps it scores higher overall, so the
aggregate ratio $s_\tau^+$ for that trajectory is large. Without
decoupling, this global preference bias leaks into every token's
gradient: the token \emph{Colosseum} in the Rome trajectory
receives a weaker learning signal not because the token-level
association ``Rome $\to$ Colosseum'' is poor, but simply because
its trajectory is globally less preferred. The residual
decomposition in Eq.~\ref{eq:fiberpo_decomp} prevents this
contamination. By subtracting the trajectory aggregate from each
token's log-ratio, the fiber gate $\tilde r_i^{\rm fiber}$
isolates the pure local association, how much ``Colosseum''
co-varies with ``Rome'' relative to what the trajectory drift
alone would predict, and gates it independently via
$\operatorname{logclip}$. Within each trajectory, token-level
learning thus operates at a uniform, unbiased scale:
$P(\text{Colosseum}\mid\text{Rome})$ and $P(\text{Eiffel
  Tower}\mid\text{Paris})$ are each refined on their own
statistical merits, free from the global preference
$P(\text{Paris trajectory}) \gg P(\text{Rome trajectory})$. The
base weight $w^{\rm base}_\tau$ then re-couples the
trajectory-level preference when the two scales are composed, so
that global significance is preserved without polluting local
precision. This is the orthogonal, non-interfering decomposition
guaranteed by the reflecting condition $\pi_{E*} \circ K =
\mathrm{id}_{\mathbf{B}}$~\cite{fiberpo2026}.

The gating map $\mathcal G$ satisfies three structural properties
established in~\cite{fiberpo2026}: (i) trajectory independence, the
Jacobian of $\mathcal G$ is block-diagonal over trajectories, fully
decoupling each trajectory's gradient, (ii) first-order agreement, at
the on-policy point ($r_\bullet = \mathbf 1$) the Jacobian reduces to
identity, recovering the true RL objective to first order near
on-policy, and (iii) scale separation, the local self-gating term has
$O(1)$ magnitude while trajectory-mediated coupling is weighted by
$1/T_\tau$, so that local gradients dominate near on-policy and
trajectory-level corrections engage only as aggregate drift grows.

We now introduce each constituent of Eq.~\ref{eq:fiberpo_decomp}.
For the base weight $w^{\rm base}_\tau$ term, the positive and
negative aggregate ratios decompose the trajectory-level drift by
sign:
\begin{equation}\label{eq:agg_ratios}
    \log s^+_\tau :=\frac{1}{T_\tau}\sum_{t=0}^{T_\tau-1} \max(\log r_{s_t(\tau),a_t(\tau)},\,0), \qquad
    \log s^-_\tau :=\frac{1}{T_\tau}\sum_{t=0}^{T_\tau-1} \max(-\log r_{s_t(\tau),a_t(\tau)},\,0).
\end{equation}

\begin{figure}[htbp]
    \centering
    \includegraphics[width=\textwidth]{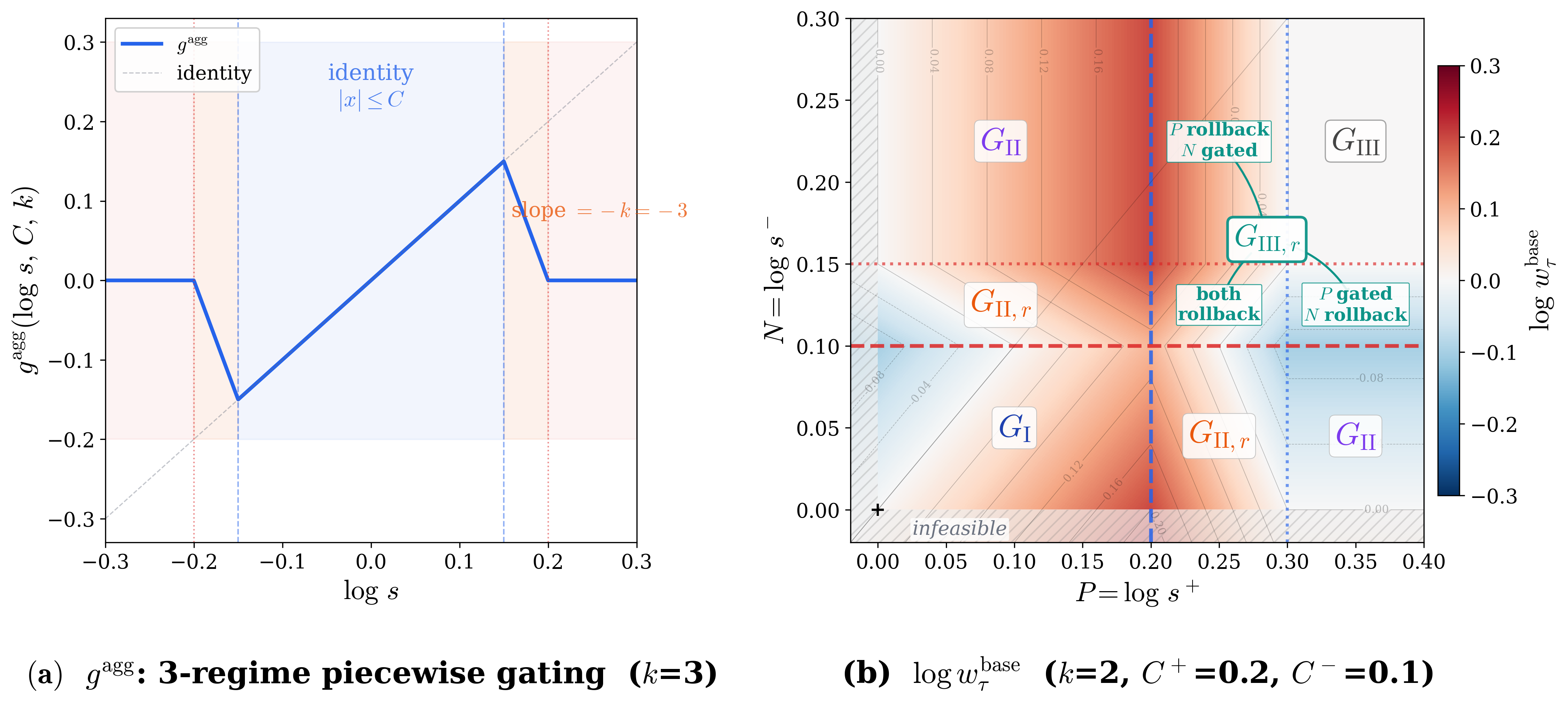}
    \caption{(a) Aggregate gate $g^{\rm agg}$ (Eq.~\ref{eq:gagg}) with three regimes: pass-through ($|x| \leq C$, slope~$1$), rollback ($C < |x| < C^* := (1+T_\tau^{-1})C$, slope~$-T_\tau$), and zeroed ($|x| \geq C^*$, output~$0$). As $T_\tau$ increases, the rollback zone narrows (width $C/T_\tau$) and $g^{\rm agg}$ approaches a hard clip at $\pm C$.
      (b) Base weight $\log w_\tau^{\rm base}$ (Eq.~\ref{eq:fiberpo_decomp}) in $(\log s^+, \log s^-)$-space with asymmetric thresholds. Dashed lines mark the budget boundaries $C^\pm$ (onset of rollback), and dotted lines mark the full-gating thresholds $C^{*\pm}$ (onset of zeroing). The five global regimes follow a non-monotonic pattern: $|\log w|$ rises through the rollback onset (G-II,r), peaks when one channel is fully gated (G-II), declines under mutual rollback (G-III,r), and collapses to zero when both channels are fully gated (G-III, $w^{\rm base}_\tau = 1$).}
    \label{fig:fiber_weight_joyai}
\end{figure}

The aggregate gating function $g^{\rm agg}$ is a piecewise-linear gate on each sign channel:
\begin{equation}\label{eq:gagg}
    g^{\rm agg}(x,C,T_\tau):= \left\{\begin{array}{ll}   x&\text{if }|x|\leq C\\[0.5em]  \operatorname{sign}(x)(T_\tau+1)C-T_\tau x & \text{if }C<|x|<(1+T_\tau^{-1})C  \\[0.5em]  0 &\text{otherwise}   \end{array}\right.
\end{equation}
where $C \in \{C^+, C^-\}$ is the per-channel trust-region budget
satisfying $C^+ + C^- = \delta$, with $C^- < C^+$ recommended to
compensate the intrinsic KL bias $\log s^-_\tau \geq \log
s^+_\tau$. The three regimes are: pass-through ($|x| \leq C$,
gate outputs $x$ unchanged), rollback ($C < |x| <
(1+T_\tau^{-1})C$, slope reverses to $-T_\tau$ producing a
restorative gradient), and zeroed ($|x| \geq (1+T_\tau^{-1})C$,
output $0$, fully blocking gradient signal). Since $w^{\rm
  base}_\tau$ is a ratio of two independently gated channels
(Eq.~\ref{eq:fiberpo_decomp}), the combined behavior produces
five global regimes (G-I through G-III) depending on which zone
each sign channel occupies. These range from nominal pass-through
(G-I: both channels transparent, base weight equals the
unmodified importance-sampling ratio) through one-channel
rollback (G-II,r: restorative gradient actively opposes the
drifting channel), one-channel fully gated (G-II: the drifting
channel is zeroed, delivering maximum one-sided correction),
mutual rollback (G-III,r), and extinction (G-III: both channels
fully gated, $w^{\rm base}_\tau = 1$, trajectory-level gradient
vanishes). The restorative rollback property is absent in PPO,
GRPO, and GSPO. Figure~\ref{fig:fiber_weight_joyai} visualizes
$g^{\rm agg}$ and the five global regimes; see~\cite{fiberpo2026}
for detailed regime definitions.

For the gated residual $\tilde r_i^{\rm fiber}$ term in
Eq.~\ref{eq:fiberpo_decomp}, the sign label $l_i$ partitions the
tokens within each trajectory into two channels, reflecting the
FBG fiber bundle structure~\cite{fiberpo2026} where the base
space $B = \mathrm{Tj}^{\theta_{\rm old}} \times \{-1,+1\}$
indexes each trajectory by a sign channel: the fiber over $(\tau,
+1)$ collects all tokens whose likelihood increased, and the
fiber over $(\tau, -1)$ collects those whose likelihood
decreased. Splitting by sign rather than averaging all log-ratios
is essential because the total trajectory drift $\overline{\log
  r}_\tau = \log s^+_\tau - \log s^-_\tau$ can be small even when
both $\log s^+_\tau$ and $\log s^-_\tau$ are individually large.
In this case, the trajectory contains many tokens that have
shifted substantially in both directions, and the
trajectory-level total variation distance ($\approx
(1/T_\tau)\sum_t |\log r_{s_t, a_t}|$) is large and may require
control. Averaging all log-ratios into a single mean would mask
this need for regulation. By tracking each sign channel
independently, $g^{\rm agg}$ detects high total variation in the
importance weights even when the signed average nearly cancels,
and can apply rollback on the offending channel without
suppressing the well-behaved one.

The log-clipping function is $\operatorname{logclip}(x,\epsilon)
:= \exp(\operatorname{clip}(\log x, \pm\epsilon))$. In ratio
space, this clamps the argument to $[e^{-\epsilon},
e^{+\epsilon}]$. The fiber residual $u_i := l_i\log r_i - \log
s_\tau^{(l_i)}$ measures each token's deviation from its
same-sign trajectory mean. Define also the opposite-sign
aggregate $v_i := -\log s_\tau^{(-l_i)}$. In terms of $u_i$ and
$v_i$, the gated residual Eq.~\ref{eq:fiberpo_decomp} can be
written equivalently as:
\begin{equation}\label{eq:gated_residual_uv}
    \tilde{r}_i^{\rm fiber} \;=\; \exp\!\bigl(\operatorname{clip}(l_i\,u_i,\;\pm\epsilon) - \operatorname{clip}(l_i\,v_i,\;\pm\epsilon)\bigr).
\end{equation}
The numerator's $\operatorname{logclip}$ acts on $e^{l_i u_i}$,
which involves only the same-sign channel aggregate
$s_\tau^{(l_i)}$, preventing opposite-channel contamination. The
denominator's $\operatorname{logclip}$ incorporates the
opposite-sign aggregate $s_\tau^{(-l_i)}$ to complete the
subtraction by the trajectory-mean log-ratio $\overline{\log
  r}_\tau$. The $\epsilon$-clip on the fiber residual $u_i$
induces three local regimes: L-I (unclipped, all tokens retain
full gradient), L-II (selective clipping of outlier tokens), and
L-III (all saturated, gradient governed entirely by the base
weight). The recommended hyper-parameter relationship $\epsilon
\ll \delta$ ensures that local regulation engages before global
regulation: the fiber gate clips outlier tokens (L-I to L-II)
well before trajectory-level aggregates reach the $g^{\rm agg}$
budget threshold (G-I to G-II). See~\cite{fiberpo2026} for the
joint local--global regime visualization on the probability simplex.

When neither $\operatorname{logclip}$ saturates (i.e., $|u_i|
\leq \epsilon$ and $|v_i| \leq \epsilon$), the clips are inactive
and we obtain $\log\tilde r_i^{\rm fiber} = \log r_i -
\overline{\log r}_\tau$, the trajectory-mean-centered log-ratio,
recovering the true linear surrogate after multiplication by the
base weight in G-I.

This fiber residual formulation yields a concrete
\emph{token-efficiency} advantage (with empirical evidence in
Section~\ref{subsubsec:single_domain}). Because the logclip acts
on $u_i = l_i\log r_i - \log s_\tau^{(l_i)}$ rather than on $\log
r_i$ directly, a token is clipped only when it deviates from the
same-sign trajectory mean by more than $\epsilon$, regardless of
the magnitude of the trajectory aggregate $\log s_\tau^{(l_i)}$
itself. Tokens that shift in concert with the trajectory-level
drift always pass through the FiberPO gating map $\mathcal{G}$
unattenuated, retaining their full gradient signal and
contributing \emph{finer, discriminative per-token update
  directions} even when the signed trajectory-level drift is
large ($|\log s_\tau^\pm| > \epsilon$). By contrast, methods such
as PPO and GRPO that clip $\log r_i$ directly tie the clip
threshold to the absolute log-ratio. Once the trajectory-level
drift exceeds the clip bound, the majority of tokens saturate
simultaneously, destroying token-level discrimination and
collapsing the gradient to a coarse trajectory-level signal.

%% file: fiberpo_exp.tex

\subsubsection{Single-Domain Evaluation}
\label{subsubsec:single_domain}

\begin{figure}[htbp]
    \centering
    \includegraphics[width=0.95\textwidth]{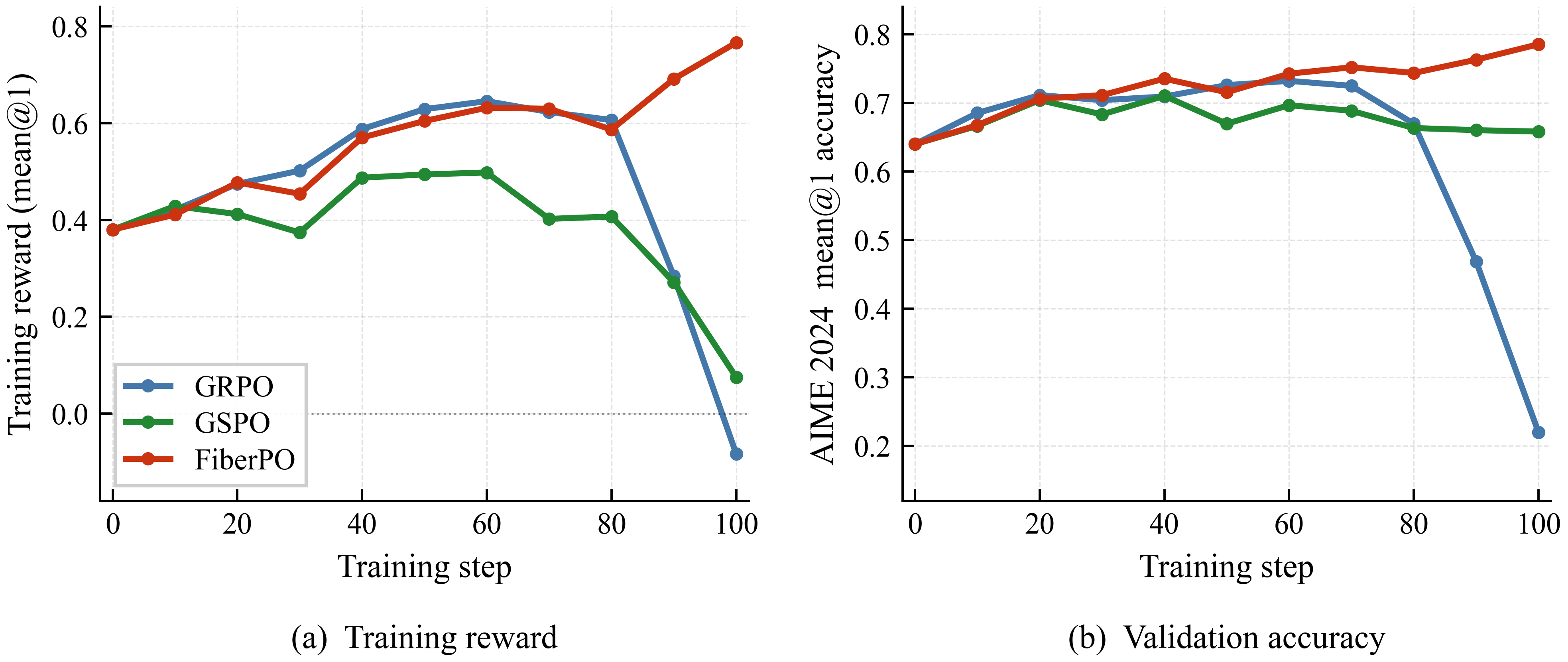}
    \caption{Single-domain RLVR on
      DAPO-Math-17k~\cite{yu2025dapo}: (a) training reward and
      (b) validation accuracy (AIME 2024 mean@1) vs.\ training
      step. GRPO collapses after step~60. GSPO stagnates. FiberPO
      improves steadily on both metrics.}
    \label{fig:math_comparison_joyai}
\end{figure}

We evaluate FiberPO in a single-domain math RLVR setting and
perform a pure algorithmic comparison against GRPO and GSPO, with
no additional stabilizers (no curriculum learning, no overlong
reward shaping or filters, etc.). We train on DAPO-Math-17k and
evaluate on AIME 2024 following the default evaluation protocol
in DAPO~\cite{yu2025dapo}. We initialize our reinforcement
learning phase using the checkpoint produced by the
aforementioned rigorous SFT and DPO stages. This presents a
deliberately challenging scenario for RL optimization, as the
extensive prior alignment significantly diminishes the policy's
entropy. Figure~\ref{fig:math_comparison_joyai} shows that
FiberPO's training and validation curves both rise monotonically
in the latter half of training. All methods are trained in
verl~\cite{sheng2024hybridflow} with matched infrastructure. We
use a learning rate of $10^{-6}$ for the RL stage.


\begin{figure}[htbp]
    \centering
    \includegraphics[width=0.95\textwidth]{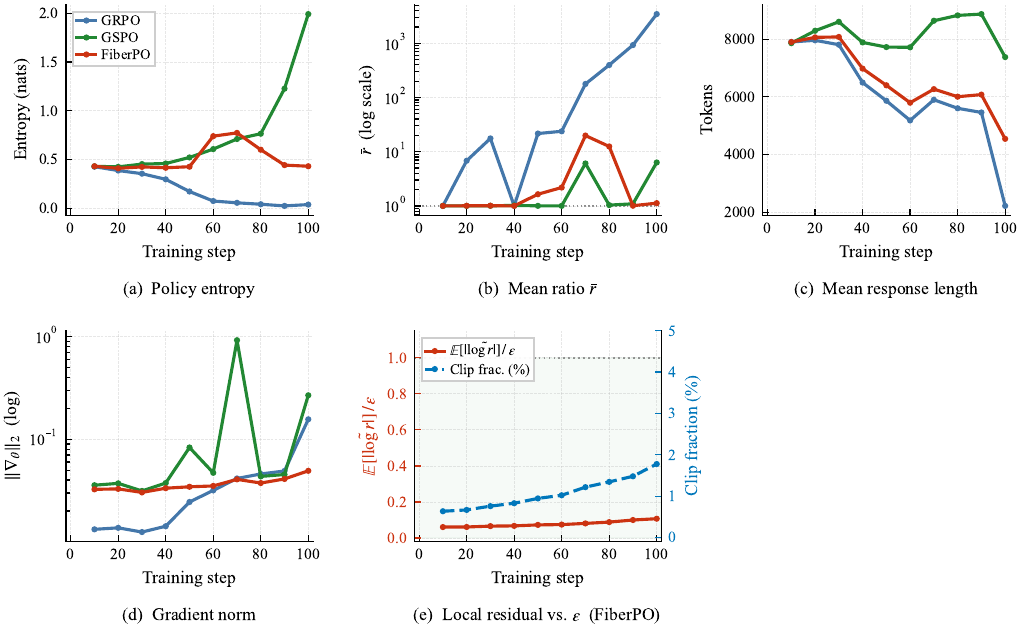}
    \caption{Training diagnostics for the single-domain DAPO math
      run. Top row (comparative, all three methods): (a) policy
      entropy, (b) mean importance ratio on log scale, (c) mean
      response length. Bottom row (FiberPO-specific): (d)
      gradient norm on log scale, (e) fiber residual and
      token-level clip fraction (both in the safe zone throughout
      training).}
    \label{fig:diagnostics_joyai}
\end{figure}

Figure~\ref{fig:diagnostics_joyai} provides training diagnostics
that can be interpreted through FiberPO's theoretical framework.
GRPO's entropy collapses to 0.038 nats (a 91\% reduction from
initialization) and its mean importance ratio exceeds $10^3$.
These observations are consistent with the structural gap
identified in Section~\ref{subsubsec:fiberpo_obj}: since GRPO
gates each token's ratio $r_i$ independently without bounding
trajectory-level aggregate drift, once the trajectory-level drift
exceeds the per-token clip bound, the majority of tokens in a
trajectory can saturate simultaneously, destroying token-level
discrimination. This plausibly triggers a feedback loop in which
imprecise updates accelerate further drift. A contributing factor
is that GRPO clips the absolute log-ratio $\log r_i$ rather than
a fiber residual: the clip threshold is shared between
trajectory-level drift and token-level variation, so trajectory
drift consumes the clip budget and forces tokens into saturation
even when their \emph{within-trajectory} deviations are small.
Additionally, GRPO lacks a restorative gradient: when a token
exceeds the clip, its gradient is zeroed rather than reversed,
providing no mechanism to oppose drift. GSPO exhibits the
complementary failure mode. Its entropy diverges to 1.99 nats with
response lengths remaining between 7{,}380 and 8{,}870 tokens:
by collapsing each trajectory to a single aggregate ratio, GSPO
suppresses within-trajectory variation and prevents the optimizer
from distinguishing token-level quality differences within the
same trajectory. The elevated entropy suggests diffuse probability
mass rather than concentration on efficient solution paths.

FiberPO addresses both failure modes through its two-scale
decomposition. It preserves entropy at 0.43 nats throughout
training while improving validation accuracy from 0.668 to 0.786.
Its mean importance ratio mostly remains at 1.13, near the
on-policy value of~1, consistent with the first-order agreement
property (first-order agreement, property~2
in~\cite{fiberpo2026}): the FiberPO Jacobian reduces to the
identity at on-policy, so each update step provides an accurate
gradient direction when the policy has not drifted far. The scale
separation property (property~3 in~\cite{fiberpo2026}) further
suggests that near on-policy, the per-token local gradient
dominates with $O(1)$ magnitude while trajectory-level
corrections contribute at most $O(1/T_\tau)$ per token, engaging
primarily only as drift grows. Unlike GRPO, the
$\operatorname{logclip}$ acts on the fiber residual $u_i =
l_i\log r_i - \log s_\tau^{(l_i)}$
(Eq.~\ref{eq:gated_residual_uv}), separating the clip budget from
trajectory-level drift so that tokens with small
within-trajectory deviations retain their full gradient signal.
The fiber residual and token-level clip fraction
(Figure~\ref{fig:diagnostics_joyai}e) remain in the safe zone
throughout training, confirming that most tokens satisfy $|u_i| <
\epsilon$ and stay in the unsaturated L-I regime. Unlike GSPO,
the per-token gated residual $\tilde r_i^{\rm fiber}$ preserves
within-trajectory discrimination, allowing the optimizer to
directly receive and update according to individual token
contributions. The gradient norm supports this interpretation:
FiberPO's norm increases only $1.5\times$ over 100 steps (0.033
to 0.049), compared to $12\times$ for GRPO and $7.5\times$ for
GSPO.

The \emph{token-efficiency property}
(Section~\ref{subsubsec:fiberpo_obj}) offers a plausible
explanation for the joint behavior of response length, entropy,
and validation accuracy. FiberPO decreases mean response length
from 7{,}902 to 4{,}543 tokens while simultaneously increasing
validation accuracy and preserving entropy. Because the
$\operatorname{logclip}$ acts on the fiber residual $u_i =
l_i\log r_i - \log s_\tau^{(l_i)}$ rather than on $\log r_i$
directly, tokens that shift in concert with trajectory-level
drift pass through unattenuated, retaining their full gradient
signal. The optimizer therefore receives discriminative per-token
directions even under moderate trajectory-level drift, which may
favor concise, correct reasoning paths over verbose ones. GRPO
also shortens responses (7{,}904 to 2{,}216 tokens at step~100),
but the accompanying entropy collapse and accuracy degradation
suggest degenerate compression rather than efficiency: once
trajectory-level drift exceeds the clip bound and token-level
discrimination is lost, the model can no longer distinguish valid
from invalid tokens. GSPO maintains high response lengths with
high variance, consistent with its suppressed per-token signal
preventing concentration of probability on efficient solution
paths.

\subsubsection{Multi-Domain Extension}
\label{subsubsec:multi_domain}

Single-domain RL training commonly degrades capabilities outside
the trained domain: a model fine-tuned exclusively on mathematics
may lose instruction-following or coding ability. Multi-domain
training addresses this by optimizing across diverse environments
simultaneously, preserving existing capabilities while gaining on
the trained domains. However, mixing heterogeneous reward
distributions intensifies the stability challenge, since
trajectory-level drift statistics vary across domains. FiberPO's
compositional two-scale gating is well suited to this setting: the
trajectory-level gate maintains per-trajectory trust regions
regardless of which domain the trajectory belongs to, while the
token-level gate preserves fine-grained credit assignment within
each domain's distinct reward structure.

We compose our training data from several domains, such as coding agent, math, knowledge, instruction following, language, and so on. All domains supply verifiable reward signals, making the full blend suitable for RLVR without learned reward models.

Following the Gaussian curriculum strategy introduced
in~\cite{nvidia2025nemotron3nanoopen}, we progressively shift
training from easier to harder prompts. Prior to training, every
prompt is profiled with $K{=}10$ rollout samples from the DPO
checkpoint to obtain an empirical pass rate $p_i \in [0,1]$.
Prompts with $p_i = 1$ (already solved) are filtered out. At
training step $t$ the target difficulty is parameterized by a
Gaussian mean
\begin{equation}\label{eq:curriculum_mean}
  \mu_t = \mu_0 + (\mu_T - \mu_0)\,\frac{t}{T},
\end{equation}
which decays linearly from $\mu_0 = 0.8$ (easy) to $\mu_T = 0.2$
(hard) over $T$ total steps
Each prompt receives a sampling weight
\begin{equation}\label{eq:curriculum_weight}
  w_i^{(t)} = \exp\!\Bigl(-\tfrac{1}{2}\bigl(\tfrac{p_i - \mu_t}{\sigma}\bigr)^{2}\Bigr),
  \qquad \sigma = 0.15,
\end{equation}
concentrating the batch around the current target difficulty.

To preserve domain balance across all heterogeneous environments,
we extend the flat Gaussian sampling with a two-level
domain-balanced scheme. At each draw, a domain group $g$ is first
selected with probability $\alpha_g$, then a prompt is drawn
within $g$ with probability proportional to $w_i^{(t)}$.
Unspecified group weights default to the square-root-proportional
heuristic $\alpha_g \propto \sqrt{N_g}$, where $N_g$ is the
number of valid prompts in group $g$, with manual overrides for
selected domains.



We reuse the training protocol and FiberPO hyperparameters from our single-domain baseline without introducing any multi-domain-specific tuning. Despite this lack of targeted adjustment, FiberPO achieves stable gains across all domains without exhibiting catastrophic degradation in any single area, confirming that the per-trajectory trust regions employed by FiberPO inherently generalize across heterogeneous reward distributions

